\documentclass[twocolumn,letterpaper,english,preprintnumbers,amsmath,amssymb,superscriptaddress,footinbib,prx]{revtex4-2}  
\usepackage{graphicx}

\usepackage{graphicx}
\usepackage{dcolumn}
\usepackage{bm}

\usepackage{xcolor} 
\usepackage[colorlinks=true, linkcolor=blue, urlcolor=blue, citecolor=blue]{hyperref}

\usepackage{graphicx}
\usepackage[inkscapelatex=false]{svg}
\usepackage{booktabs}

\usepackage{amssymb,amsthm,amsmath}
\usepackage{xcolor,paralist,hyperref,titlesec,fancyhdr,etoolbox}
\usepackage{tikz}
\usepackage{tikz-qtree,tikz-qtree-compat}
\usepackage{cleveref}
\usepackage{float}     

\usepackage{tcolorbox} 
\usepackage{listings}  
\usepackage{xcolor}    

\tcbuselibrary{breakable,skins,listings}

\usepackage[commandnameprefix=always]{changes}
\definechangesauthor[name={Doron Sivan}, color=blue]{ds}

\newtcolorbox[auto counter]{prompt}[2][]{
  label = #1,
  title={Prompt~\thetcbcounter\quad#2},
  breakable,
  enhanced,
  colback=white,
  listing only,
  listing options={
    basicstyle=\ttfamily,
    breaklines=true,
    language=Tex,
    escapechar=§,
  },
}

\lstdefinelanguage{json}{
  basicstyle=\normalfont\ttfamily,
  numbers=left,
  stepnumber=1,
  numbersep=5pt,
  showstringspaces=false,
  breaklines=true,
  frame=single,
  backgroundcolor=\color{white},
  literate=
   *{0}{{{\color{blue}0}}}{1}
    {1}{{{\color{blue}1}}}{1}
    {2}{{{\color{blue}2}}}{1}
    {3}{{{\color{blue}3}}}{1}
    {4}{{{\color{blue}4}}}{1}
    {5}{{{\color{blue}5}}}{1}
    {6}{{{\color{blue}6}}}{1}
    {7}{{{\color{blue}7}}}{1}
    {8}{{{\color{blue}8}}}{1}
    {9}{{{\color{blue}9}}}{1}
    {:}{{{\color{red}{:}}}}{1}
    {,}{{{\color{red}{,}}}}{1}
    {\{}{{{\color{black}{\{}}}}{1}
    {\}}{{{\color{black}{\}}}}}{1}
    {[}{{{\color{black}{[}}}}{1}
    {]}{{{\color{black}{]}}}}{1},
}

\newtcolorbox{userinput}{
  breakable,
  enhanced,
  colback=blue!5,
  colframe=blue!75!black,
  title=User Input,
}

\usepackage[table]{xcolor} 

\hypersetup{ colorlinks=true, linkcolor=black, filecolor=black, urlcolor=black }

\renewcommand{\sectionmark}[1]{}
\renewcommand{\subsectionmark}[1]{}

\begin{document}

\title{Semantic Chunking and the Entropy of Natural Language}

\newcommand{\ourtitle}{Semantic Chunking and the Entropy of Natural Language}

\author{Weishun Zhong}
\email{weishunzhong.ai@gmail.com}
\affiliation{School of Natural Sciences, Institute for Advanced Study, Princeton, NJ, 08540, USA}

\author{Doron Sivan}
 \affiliation{Department of Brain Sciences,
Weizmann Institute of Science, Rehovot, 76100, Israel}

\author{Tankut Can}
\affiliation{Department of Physics, Emory University, Atlanta, GA 30322, USA}

\author{Mikhail Katkov}
 \affiliation{School of Natural Sciences, Institute for Advanced Study, Princeton, NJ, 08540, USA}
 \affiliation{Department of Brain Sciences,
Weizmann Institute of Science, Rehovot, 76100, Israel}

\author{Misha Tsodyks}
\email{mtsodyks@gmail.com}
 \affiliation{School of Natural Sciences, Institute for Advanced Study, Princeton, NJ, 08540, USA}
 \affiliation{Department of Brain Sciences,
Weizmann Institute of Science, Rehovot, 76100, Israel}

\begin{abstract}

Humans and large language models can predict next letter or word from its prior context much better than random guessing, indicating strong redundancy of language viewed as a stochastic process. Quantitatively this redundancy was estimated by Shannon to be around 80\%, which means that every letter of a printed English text conveys approximately 1 bit of information and not 4.8 bits that 27 letters (including spaces) could potentially carry. This estimate was later confirmed by using autoregressive token probabilies computed by large language models. However, the statistical organization of language that give rise to such a large redundancy remains unclear. Here we introduce a statistical framework of language linking its redundancy to the hierarchical semantic organization of text. To this end, we use large language models to recursively segment any given text into semantically coherent chunks, inducing a ``semantic tree'' that spans the whole range of text organization, beginning from its main idea to individual tokens (words). For a large corpus of texts of a particular type, say fiction stories, the resulting ensemble of semantic trees is characterized by specific statistical regularities, giving rise to a ``structural'' entropy rate defined in this study. Surprisingly, we discovered that for several datasets considered in this work, semantic tree entropy rate was quite close to LLM-measured quantity and exhibited a similar trend across corpus. In particular, simpler texts like children stories exhibit lower branching in their semantic trees and correspondingly lower entropy rates, whereas fiction and poetry exhibit progressively larger branching factors and greater entropy rates. These results suggest that hierarchical semantic organization of texts is an important factor in their overall information transmission rates. 

\end{abstract}

\maketitle


\section{Introduction}

Understanding a natural text is closely related to the ability to predict what comes next, i.e. reducing uncertainty about the text's continuation. Historically, this uncertainty has been quantified by the entropy rate of language, viewed as a stochastic process, in terms of an average information carried by each character or word. The first estimate of the entropy rate of printed English was obtained by Shannon using the now-famous ``guessing'' game procedure, where a subject was trying to guess the next character after reading passages of various lengths \cite{shannon1951prediction, cover1978convergent}. The resulting estimate of approximately 1 bit per character, which implies 80\% redundancy relative to the 5 bits per character of random text, and similar estimates have since been obtained across languages \cite{barnard1955statistical,newman1960redundancy}. Modern auto-regressive large language models (LLMs) are trained to predict the next token (the basic unit of text, such as a word, part of a word, or punctuation mark) from the preceding context \cite{bengio2003neural, radford2018improving}. At inference time, these models can be used to estimate language entropy rates from token probabilities, yielding values comparable to those obtained from Shannon-style guessing games \cite{ho2024algorithmic,takahashi2018cross,kaplan2020scaling,valmeekam2023llmzip}. There is currently no solid theoretical understanding of the statistical regularities of language that contribute to its empirically observed redundancy values. 

In the current study, we focus on the hierarchical nature of semantic organization of natural texts: words in a sentence have to satisfy the grammar, sentences have to relate to each other in a particular way to convey particular events, and the end of the story has to correspond to how the story begins. This hierarchical organization of texts generates statistical redundancies at all levels. A reasonable, if necessarily simplified, way to capture this organization is via ``semantic trees,'', i.e. a set of keypoints of various degrees of abstraction, with more abstract keypoints elaborated with a number of less abstract ones. We further assume that each keypoint summarizes a particular span of text, which can itself be segmented into shorter spans that capture progressively more detailed information
\cite{kintsch1978toward,mann1988rhetorical, taboada2006rhetorical, zhong2025random}. At the very top of the tree there is the most general idea, or keypoint, characterizing the whole text; at the next layer, the text is segmented into several coherent ``chunks'' communicating a set of keypoints that together elaborate the main keypoint of the text, etc. This recursive procedure results in a semantic tree that is unique to each text (and possibly to each reader that may comprehend the given text differently from other readers). In this representation, a text is characterized not only by its surface sequence of tokens, but also by a hierarchical structure that constrains which continuations are plausible at each scale. If these constraints account for a substantial part of linguistic predictability, then the semantic tree structure should be closely related to the entropy rate measured by next-token prediction. 

\begin{figure*}
    \centering
        \includegraphics[width=1.0\textwidth]{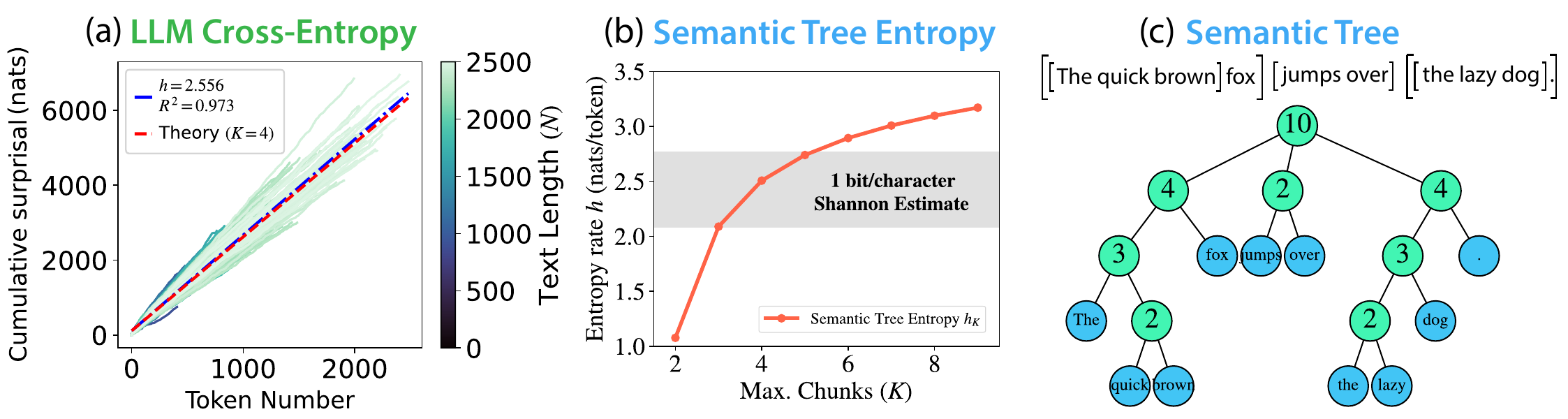}
        \caption{\textbf{Semantic tree entropy contributes significantly to the overall predictability of text.} \\
        \textbf{(a)} Cumulative LLM surprisal, \(H_N = -\sum_{i=1}^N \log P(t_i|t_{<i})\), estimated using \texttt{Llama-3.3-70B} \cite{grattafiori2024llama3herd}) for 925 fiction stories from \texttt{RedditStories}~\cite{fan2018hierarchical}. Color indicates text length \(N\). The blue dash-dotted line is a linear fit whose slope gives the LLM-based entropy-rate estimate \(h_{\mathrm{LLM}}\). Slope of the red dashed line correspond to the entropy computed from the random tree ensemble at $K=4$. Intercept of the red dashed line is the same as the blue dashed-dotted line. 
        \textbf{(b)} Entropy rate \(h_K\) of the random tree ensemble, as an approximation for the semantic trees, versus the maximum number of semantic chunks \(K\) per level. For \(K=4\), corresponding to the typical human working memory (WM) capacity, the predicted entropy rate lies within the range of Shannon's estimate of approximately 1 bit per character for printed English (shaded band). This scale agreement motivates the hypothesis that hierarchical semantic organization contributes substantially to language entropy.
        \textbf{(c)} The chunking procedure applies recursive semantic segmentation on a text until reaching the single-token level, thereby parsing the document into a hierarchical tree of spans whose leaves are tokens (a ``semantic tree''). }
\label{fig:entropy_curve}
\end{figure*}

In our recent work \cite{zhong2025random}, we argued that when all the texts of a particular type are considered, the partition of text into semantic chunks can be described as a random process, and the corresponding ensemble of trees can be approximated by the one where every intermediate chunk of text is segmented randomly into smaller pieces by placing a certain number of divisions at random positions between the tokens. The largest number of resulting text chunks (denoted by $K$ throughout the paper) is the branching factor of the tree ensemble and a parameter of the model. In the context of narrative memory, we further speculate that this parameter represents the maximum number of keypoints that can be simultaneously maintained during comprehension, and is determined by human working memory capacity.

The random tree ensemble \cite{zhong2025random} provides a statistical description of the hierarchical semantic organization of text. A natural question is therefore how much entropy is associated with this structure and how it compares with standard estimates of language entropy. In the present work, we further develop the tree model of meaningful texts and establish a connection between hierarchical semantic structure and language entropy. We begin by calculating the entropy rate of the random tree ensemble and comparing it with Shannon's estimate for printed English. We then use LLMs to recursively segment a collection of fiction stories into empirical semantic trees and examine the extent to which their aggregate statistics are captured by the random tree ensemble. Finally, we extend the analysis to other types of text, including children's stories and modern poetry, to test whether differences in entropy rate reflect systematic differences in hierarchical semantic structure.

\section{Entropy rate of English and LLM perplexity}
\label{sec:llm_entropy}

In this section we recapitulate the entropy rate estimation using large language models. Entropy rate of language is defined as the uncertainty associated with the next token given its preceding context \cite{shannon1951prediction}. Auto-regressive large language models (LLMs) \cite{vaswani2017attention,radford2018improving} are trained to approximate this conditional distribution by minimizing cross-entropy loss on observed text \cite{bengio2003neural}. At inference time, rather than using an LLM for generation, we use it as a measurement device: given a text, the model assigns probabilities to the observed tokens, allowing us to estimate their average surprisal. Concretely, for a text represented as a length-\(N\) token sequence \(\{t_i\}_{i=1}^N\), we evaluate the model left-to-right by conditioning on the prefix \(t_{<i}\) and recording the probability \(P(t_i | t_{<i})\) assigned to the observed token \(t_i\). The per-token surprisal is therefore \(-\log P(t_i | t_{<i})\), and its average defines the LLM-based cross-entropy rate estimate for a given text:
\begin{equation}
\label{eq:entropy_llm}
    h_{\mathrm{LLM}} = -\frac{1}{N}\sum_{i=1}^N \log P(t_i|t_{<i}) .
\end{equation}
By standard information-theoretic arguments, the cross-entropy upper-bounds the true entropy rate of the text distribution when the model distribution differs from the true distribution \cite{cover1978convergent,brown1992estimate}. Equivalently, Eq.~\eqref{eq:entropy_llm} is the log-perplexity of the model on the text, \(h_{\mathrm{LLM}}=\log \mathrm{PPL}\) \cite{jelinek1977perplexity,chen1999empirical,huggingface_transformers_perplexity}.

To estimate a corpus-level entropy rate, we applied Eq.~\eqref{eq:entropy_llm} to 925 fiction stories randomly sampled from the \texttt{RedditStories} dataset \cite{fan2018hierarchical}. We computed the cumulative surprisal of each text, $H_N = -\sum_{i=1}^N \log P(t_i|t_{<i})$, and fitted its dependence on text length \(N\) across the sample. The fitted slope gives the mean per-token cross-entropy rate, \(h_{\mathrm{LLM}} = 2.56\) nats/token (blue dash-dotted line in Fig.~\ref{fig:entropy_curve}(a)). Since modern tokenizers encode English text at roughly 3--4 characters per token \cite{limisiewicz2023tokenization}, this value is consistent with Shannon's estimate of 1 bit per character.

\section{Theory of the random tree ensemble}

In the rest of the paper, we will evaluate the relation between the entropy rate of natural texts and the statistics of their hierarchical semantic organization. To this end, we start by describing a statistical theory of random tree ensemble, where we first layout the general theory and then show that it is captures the hierarchical semantic structure of real texts. Following \cite{zhong2025random}, we consider an ensemble of trees obtained with a self-similar segmentation of texts \cite{bertoin2006fragmentation,stanley2012enumerative1,zhong2025random}. In particular, a text of $N$ tokens is divided into $K$ (possibly empty) chunks by placing $K-1$ divisions randomly in the $N+K-1$ positions of the divisions and the $N$ tokens. When two or more divisions are adjacent, the corresponding chunk(s) will be empty; and if all the divisions fall either before the first token or after the last token, the chunk is not split. The same procedure is then applied recursively to each chunk of more than one token, until each remaining chunk is either of a size one or does not split. Mathematically, this construction defines a one-parameter family (indexed by \(K\)) of \textit{random tree ensemble}s, where every chunk of size $n$ can be split in $Z_K(n) = \binom{n+K-1}{K-1}$ different ways that we consider to be all equally likely. We will postpone the discussion of how well the ensemble introduced above corresponds to the semantic organization of real texts to the next session. The full mathematical analysis of the random tree ensemble will be presented in a separate publication \cite{sivan2026}; here we present the main results.

Because splitting of different chunks is independent on each other, every possible tree ($T$) in the ensemble with chunk sizes $\{n_i\}$ ($i\in$ internal nodes of $T$ as the leaves do not split further), is assigned a probability 

\begin{equation}
\label{eq:tree_log_prob}
P(T) = \prod_i \frac{1}{Z_K(n_i)}.
\end{equation}
and one can define the entropy of the random tree ensemble including all possible trees of a given size $N$ (i.e. the size of the root). Since we assume that this statistical ensemble approximates the ensemble of all texts with $N$ tokens, it is instructive to compare the entropies of these ensembles. To calculate the entropy of the random tree ensemble, we use the following recursive formula:
\begin{align}
\label{eq:H_recursive}
    H(N) = \ln Z_K(N) + \frac{1}{Z_K(N)} \sum_{ {n_i}:\sum_{i=1}^{i=K} n_i = N} \sum_{i=1}^{i=K} H(n_i),
\end{align}
with $H(1)=0$. The first term in this formula is the entropy of the initial splitting of the text into $K$ chunks, while the second term is the conditional entropy of the tree ensemble conditioned on the first split which is written as an average over all possible divisions of $N$ into $K$ non-negative numbers $n_i \ge 0; i=1,...,K$, where we used the assumption that all divisions have equal probabilities (see \cite{cover1999elements}). We can further simplify the above equation by noting that for the splitting model considered above, each of $K$ chunks has the same probability distribution
\begin{align}
\label{eq:chunk_distribution}
P(n|N) = \frac{Z_{K-1}(N-n)}{Z_K(N)},
\end{align}
e.g., for the first chunk $P(n|N)$ is obtained by counting all the configurations where the first chunk has $n$ tokens (see \cite{zhong2025random} for more details). Recursive equation Eq.~\eqref{eq:H_recursive} can then be rewritten in a more concise form as (see SI):
\begin{align}
\label{eq:H_LHS_main}
    H(N) = \ln Z_K(N) + K \sum_{n=2}^{N-1} \frac{Z_{K-1}(N-n)}{Z_K(N)} H(n).
\end{align}
In \cite{sivan2026} we show rigorously that for large trees, the entropy of the corresponding ensemble scales linearly with their size $N$:
\begin{align}
\label{eq:entropy_extensive}
    H(N) \simeq h_K N .
\end{align}
where \(h_K\), which we call ``entropy rate'' in analogy to stochastic processes, depends only on \(K\) and can be calculated analytically (see \cite{sivan2026}) or obtained by numerical iteration of equation Eq.~\eqref{eq:H_LHS_main} (see SI for instructive examples for proving the special case of $K=2$ and $K \gg 1$). As shown in Fig.~\ref{fig:entropy_curve}(b), $h_K$ increases monotonically with $K$. Surprisingly, for $K=4$ the value of $h_K \sim 2.5$ nats/token is close to the entropy rate obtained for fiction stories using LLMs (red dashed line in Fig.~\ref{fig:entropy_curve}(a)). Since we hypothesized in \cite{zhong2025random} that $K=4$ is the correct branching factor for typical English texts, this result suggests that tree entropy rate could be an important (or even dominant) contribution to total entropy rate. To further elaborate on this hypothesis, in the following we use fiction as a testbed and perform LLM-assisted semantic segmentation on a large collection of stories to study its properties. 

\section{Semantic chunking}
\label{sec:semantic_chunking}

\begin{figure*}
    \centering
        \includegraphics[width=0.9\textwidth]{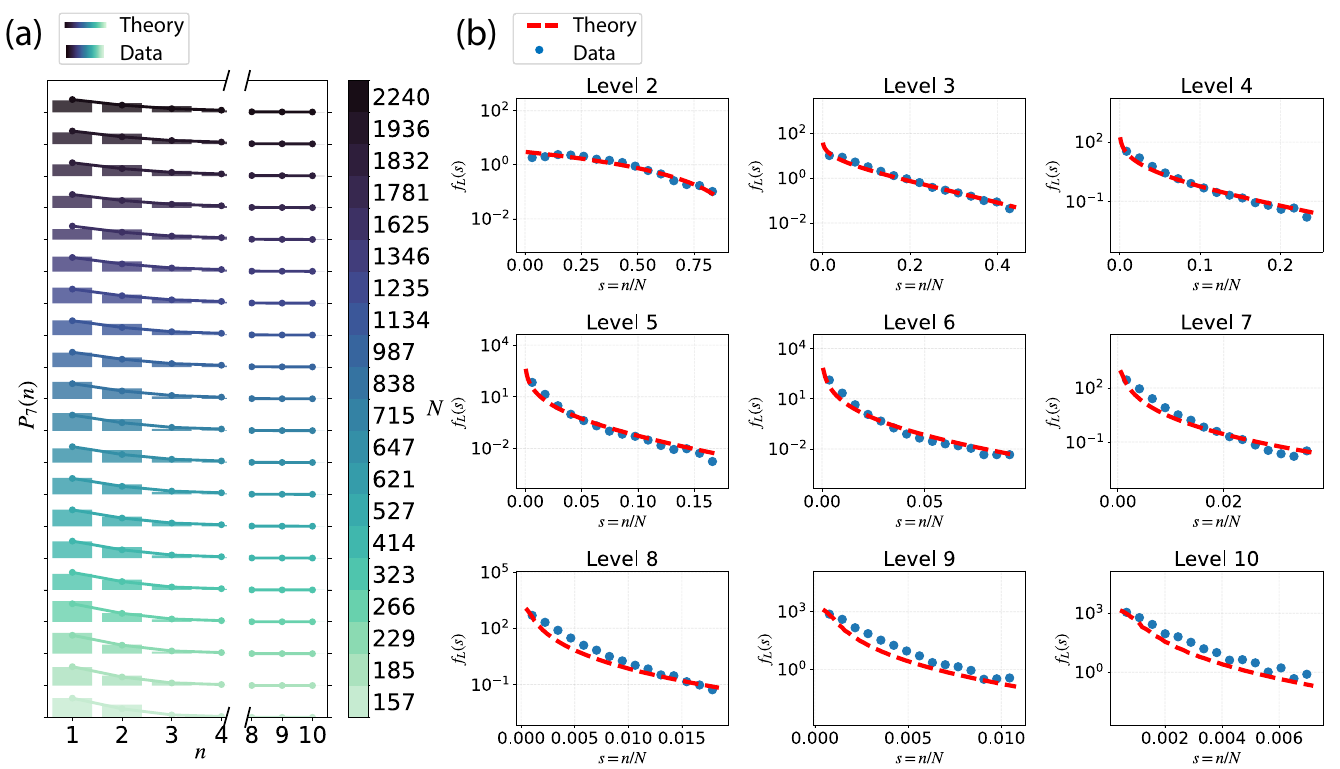}
        \caption{\textbf{Random \(K\)-ary trees capture the chunk-size statistics of empirical semantic trees.} \\
        \textbf{(a)} Empirical chunk-size distributions at level \(L=7\) for 20 fiction stories from \texttt{RedditStories}~\cite{fan2018hierarchical}, compared with theoretical distributions \(P_L(n|N)\) computed from the random tree model for each text length \(N\). 
        \textbf{(b)} After normalizing chunk sizes by the root length \(N\), empirical distributions pooled across 925 texts from \texttt{RedditStories} collapse onto the large-\(N\) scaling predictions \(f_L\) at multiple levels \(L\).}
\label{fig:tree_statistics}
\end{figure*}

In this section we study the hierarchical structure of natural texts and compare it to the random tree ensemble introduced above. To this end we use an LLM to recursively identify semantically coherent chunks at multiple scales, thereby inducing a hierarchical decomposition of each text into a semantic tree \cite{qu2025semantic,kiss2025max,duarte2024lumberchunker}. Starting from the full document, each text span that contains more than one token is passed to an LLM with a system prompt instructing it to segment the span into at most \(K\) contiguous, semantically coherent chunks (see SI for the full prompt). If the LLM returns more than one chunk, these chunks become the children of the current node, and the same procedure is recursively applied to each child containing more than one token. If the LLM returns a complete chunk containing multiple tokens, or if a span contains only one token, that span is accepted as a leaf and exits the chunking procedure. The result is a hierarchical tree of text spans, in which individual tokens (or small chunks of tokens) appear as leaves and internal nodes correspond to coherent spans at progressively coarser resolutions (Fig.~\ref{fig:entropy_curve}(c)).

We apply our semantic chunking algorithm to 925 fiction stories randomly sampled from the \texttt{RedditStories} dataset, a collection of human-written stories from the online forum Reddit \cite{fan2018hierarchical}, inducing one semantic tree per story. The algorithm has a single parameter, the maximum branching factor \(K\). The sampled stories range from \(\sim 100\) to \(\sim 2500\) tokens. All resulting semantic trees have at least 7 levels of depth, and the longest stories reach depths of up to \(\sim 12\) levels. We can then compare the empirical chunk size distributions at different layers with those predicted by the random tree ensemble described in the previous section. For the first layer, the distribution $P_1(n|N)$ is given in Eq.~\eqref{eq:chunk_distribution}, for all other layers \(P_L(n|N)\) is calculated in \cite{zhong2025random} (see SI for a recap). As an illustrative example, solid curves in Fig.~\ref{fig:tree_statistics}(a) plot \(P_L(n|N)\) for different values of \(N\) at an intermediate level (\(L=7\)), and the histograms are the corresponding empirical chunk-size distributions from a random subset of 20 semantic trees.

\begin{table*}[t]
\caption{Goodness-of-fit, measured by the average KL divergence \(\langle D_{\mathrm{KL}}\rangle\), across corpora \cite{eldan2023tinystories, richardson2013mctest, fan2018hierarchical, hf_suayptalha_poetry_foundation_poems} for different values of \(K\). For each corpus and model (\texttt{Llama-4-Maverick} \cite{meta2025llama4herd} and \texttt{DeepSeek-V3.1} \cite{liu2024deepseek}), the optimal $K^\star = \arg\min_{K} \langle D_{\mathrm{KL}}\rangle$ is highlighted in green.}
\label{table:kl_div}
\begin{ruledtabular}
\begin{tabular}{l*{10}{c}}
\textbf{$K$}
& \multicolumn{2}{c}{\textbf{\begin{tabular}{c}
AI-Generated \\Children's Stories\\
(\texttt{TinyStories}~\cite{eldan2023tinystories})
\end{tabular}}}
& \multicolumn{2}{c}{\textbf{\begin{tabular}{c}
Human-Written \\Children's Stories\\
(\texttt{MCTest}~\cite{richardson2013mctest})
\end{tabular}}}
& \multicolumn{2}{c}{\textbf{\begin{tabular}{c}
Fiction \\
(\texttt{RedditStories}~\cite{fan2018hierarchical})
\end{tabular}}}
& \multicolumn{2}{c}{\textbf{\begin{tabular}{c}
Modern Poetry \\
\Big(\begin{tabular}{@{}c@{}}
\texttt{PoetryFoun-} \\
\texttt{dationPoems}
\end{tabular}~\cite{hf_suayptalha_poetry_foundation_poems}\Big)
\end{tabular}}}
\\
\\
\cline{2-3}
\cline{4-5}
\cline{6-7}
\cline{8-9}
\cline{10-11}
&
\textbf{Llama} & \textbf{DeepSeek}
& \textbf{Llama} & \textbf{DeepSeek}
& \textbf{Llama} & \textbf{DeepSeek}
& \textbf{Llama} & \textbf{DeepSeek}
\\
\hline
$2$
& \cellcolor{green!25}0.144  & \cellcolor{green!25}0.178
& 0.360 & 0.437
& 0.12 & 0.106
& N/A & N/A
\\
$3$
& 0.160 & 0.191
& \cellcolor{green!25}0.128 & \cellcolor{green!25}0.183
& 0.121 & 0.103
& N/A & N/A
\\
$4$
& 0.164 & 0.268
& 0.164 & 0.194
& \cellcolor{green!25}0.0542 & \cellcolor{green!25}0.0493
& 0.323 & 0.281
\\
$5$
& 0.306 & 0.369
& 0.199 & 0.218
& 0.111 & 0.165
& 0.172 & 0.266
\\
$6$
& 0.313 & 0.813
& 0.324 & 0.245
& 0.167 & 0.243
& \cellcolor{green!25}0.149 & \cellcolor{green!25}0.135
\\
$7$
& N/A & N/A
& N/A & N/A
& N/A & N/A
& 0.174 & 0.246
\\
$8$
& N/A & N/A
& N/A & N/A
& N/A & N/A
& 0.307 & 0.288
\\
\end{tabular}
\end{ruledtabular}
\end{table*}

Since stories have different lengths \(N\), however, chunk sizes cannot be pooled directly. To overcome this difficulty, we used the asymptotic recursive expression for chunk size distributions $P_L(n)$ at successive levels $L$ that is valid for large trees ($N \gg1$), derived in \cite{zhong2025random} and reproduced in the SI:
\begin{align}
    P_L(n) & \simeq \frac{1}{N} f_L(n/N), \label{eq:asymp_Pn} \\
    f_{L}(s) &=
    \begin{cases}
        \displaystyle \int_s^1 \frac{dr}{r}\, (K-1)(1-r)^{K-2}\, f_{L-1}\!\left(\frac{s}{r}\right), & L>1,\\[2.2ex]
        \displaystyle \delta(s-1), & L=1.
    \end{cases}
\end{align}
Using the scaling function \(f_L\) that only depends on $K$ but not on $N$, we can remove the explicit dependence on text length by normalizing chunk sizes as \(s_L := n/N\). At e level \(L\),  chunks from different stories can then be pooled and their normalized sizes treated as samples from a single empirical distribution \(\hat{f}_L\).  We divide the range between the minimum and maximum observed values of \(s_L\) into 15 equally spaced bins. This procedure enables a direct comparison between \(\hat{f}_L\) and the theoretical scaling function \(f_L\) using all chunks from the 925 semantic trees at each level, with both functions depending on the choice of $K$. In Fig.~\ref{fig:tree_statistics}(b) these functions are computed for $K=4$, a value suggested in \cite{zhong2025random} for narrative texts, showing a reasonable agreement across levels.  The remaining errors indicate the deviation of semantic chunking from the random splitting model and also result from finite-size and finite-sample effects, see SI for a more detailed discussion. 

To quantify goodness of fit, we measured the discrepancy between empirical and theoretical distributions using the KL divergence, averaged across tree levels:
\begin{align}
    \langle D_{\mathrm{KL}}(\text{data}\,\|\,\text{theory}) \rangle
    = \frac{1}{L}\sum_{l=1}^{L}\sum_{s \in \mathcal{S}_l}\hat{f}_{l}(s)\log\frac{\hat{f}_{l}(s)}{f_{l}(s)} ,
\end{align}
where \(\mathcal{S}_l\) denotes the set of bins for normalized chunk sizes at level \(l\), and \(\hat{f}_{l}\) and \(f_{l}\) are the empirical and theoretical scaling functions, respectively. 

To find the value of $K$ for which the semantic trees most resemble the random chunking ensemble, we repeated the above procedure for all $K$s in the range of $2$ to $8$, using a random subset of $50$ stories from our data set. Indeed, we found that $K^\star = 4$ minimizes this measure (see Table~\ref{table:kl_div} third column), agreeing with the choice of $K=4$ in \cite{zhong2025random}. 

To better assess the quality of the agreement between theory and data, we perform explicit random chunking on the same texts used in Table~\ref{table:kl_div}, following the random tree model of \cite{zhong2025random}. The resulting KL divergence, \(0.031\), provides a finite-size baseline for the best achievable fit to the theoretical scaling form. For semantic chunking at \(K=4\), the corresponding KL divergence is \(0.054\), which is comparable to this baseline. We therefore wondered whether semantic segmentation performed by LLM indeed identifies coherent chunks of a text. To this end, we show in the SI that the two segmentations can still be distinguished by their preservation of linguistic structure: semantic chunking recovers substantially more linguistic clauses than random chunking, despite receiving no supervision from clause annotations. Thus, although the aggregate chunk-size statistics of semantic trees are similar to random tree ensemble, semantic chunking preserves contextual semantic structure that is absent in random partitions of text.

\section{Typicality of the tree entropy}

\begin{figure*}[t!] 
    \centering
        \includegraphics[width=0.95\textwidth]{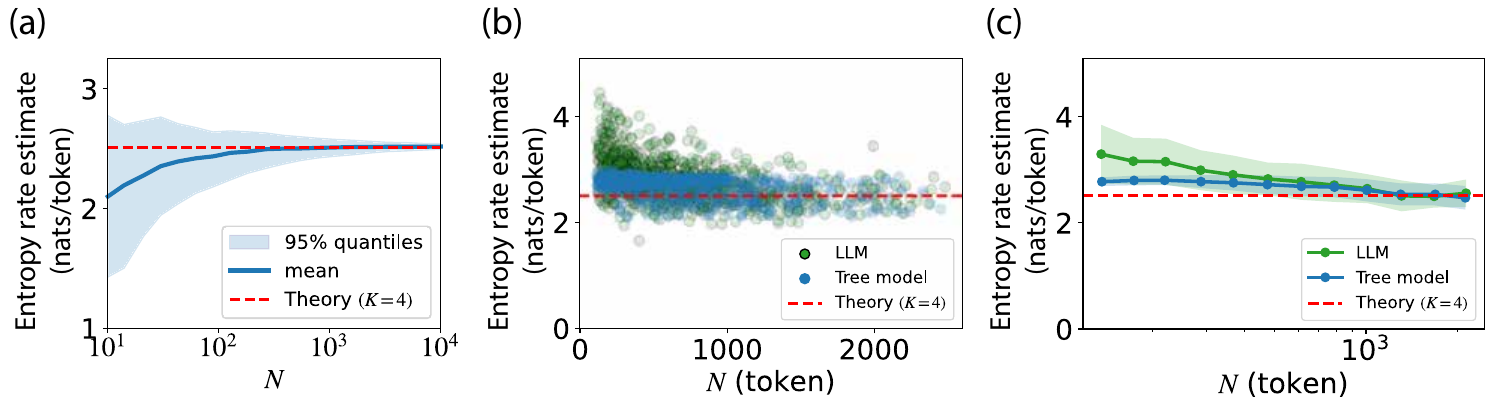}
        \caption{\textbf{Fluctuation vanishes asymptotically for both LLM- and tree-based estimates of text entropy.} \\  
        \textbf{(a)} Entropy-rate estimates from individual random tree realizations (simulated using Eq.~\eqref{eq:chunk_distribution}) concentrate around the predicted value as \(N\) increases, consistent with the emergence of typical trees.
        \textbf{(b)} Per-token entropy estimates for 925 fiction stories from \texttt{RedditStories} \cite{fan2018hierarchical}, computed in two ways: from LLM perplexity (green dots) and from the likelihood of empirical semantic trees obtained via chunking (blue dots). As \(N\) increases, both estimates fluctuate around the predicted entropy rate. 
        \textbf{(c)} Data from (b) binned by text length ($N$) uniformly on a log-scale. Solid curves indicate the mean within each bin, and the shaded regions indicate one standard deviation.  
        }
\label{fig:tree_entropy}
\end{figure*}

The fiction corpus shows substantial variability in cumulative surprisals across individual texts (e.g., the spread of traces in Fig.~\ref{fig:entropy_curve}(a)). For an entropy rate to be a meaningful characterization of the ensemble, this variability should become negligible in the large-\(N\) limit. Indeed, the random tree ensemble exhibits an \textit{Asymptotic Equipartition Property} \cite{sivan2026}: in the limit of large \(N\), the normalized negative log-likelihood of a tree converges in probability to the entropy rate:
\begin{align}
\label{eq:typicality}
-\frac{1}{N}\log P(T) \xrightarrow{N\to\infty} h_K,
\end{align}
i.e. for large $N$ most of the trees generated according to the above procedure belong to a set of ``typical trees'' with per-leaf log-likelihoods concentrated around entropy rate $h_K$; the total probability of this set approaches $1$ in this limit.  This concentration is illustrated in Fig.~\ref{fig:tree_entropy}(a): simulations at increasing \(N\) show that realization-to-realization fluctuations in the entropy rate estimate vanish, converging to the theoretical value \(h_K \approx 2.5\) nats/token for \(K=4\).

Additionally, we can compare the changed to entropy rate estimate from individual texts within the same corpus in two complementary ways: (i) from LLM perplexity (Eq.~\eqref{eq:entropy_llm}); and (ii) from the log-probability of observing the empirical semantic tree under the random tree ensemble evaluated at the corpus-level inferred value \(K^\star\) (Eq.~\eqref{eq:tree_log_prob}). For the fiction corpus, the entropy rates of 925 stories estimated using these two approaches are shown in Fig.~\ref{fig:tree_entropy}(b). As \(N\) increases, both estimates approach the predicted entropy rate \(h_{K^\star}\) (red dashed line), while their fluctuations decrease asymptotically (Fig.~\ref{fig:tree_entropy}(c)), consistent with the typicality property in Eq.~\eqref{eq:typicality}.

\section{Other types of texts}

The above results show that typical English texts, such as fiction stories, exhibit hierarchical semantic organization comparable to random tree ensemble with branching factor \(K=4\), and that, surprisingly, LLM-evaluated entropy rate of texts is similar to mathematically computed entropy rate of the random tree ensemble. In this section, we look at other datasets in order to study the generality of the above observations. In particular, we considered three additional datasets: \texttt{TinyStories}, a corpus of short AI-generated children's stories written with simple vocabulary and grammar \cite{eldan2023tinystories}; \texttt{MCTest}, a set of short human-written children's stories designed for machine-comprehension tasks \cite{richardson2013mctest}; and \texttt{PoetryFoundationPoems}, a collection of modern poems from the Poetry Foundation \cite{hf_suayptalha_poetry_foundation_poems}. 

\begin{figure}
    \centering
        \includegraphics[width=0.3\textwidth]{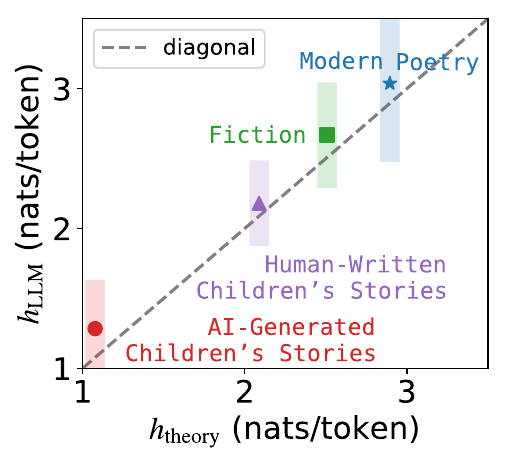}
        \caption{\textbf{Semantic tree entropy predicts entropy rates across corpora.} Entropy rate of different corpora estimated in two ways: theoretical values correspond to \(h_K\) (Eq.~\eqref{eq:entropy_extensive}) evaluated at optimal \(K\) for each corpus (Table~\ref{table:kl_div});  \(h_{\mathrm{LLM}}\) are obtained from cross-entropy estimate (Eq.~\eqref{eq:entropy_llm}) on each corpus.  }
\label{fig:entropy_corpus}
\end{figure}

For each dataset, we repeated the analysis on 50 randomly selected texts. We found that different corpora are characterized by different optimal branching factors \(K^\star\) (Table~\ref{table:kl_div}), with the smallest \(K^\star\) for AI-generated children's stories and the largest for modern poetry. A corresponding trend is observed in the LLM-estimated entropy rates obtained from the regression slopes for each dataset. Moreover, across all four corpora, the empirical entropy rates are close to the theoretical predictions of the random tree ensemble evaluated at the corresponding \(K^\star\) values (Fig.~\ref{fig:entropy_corpus}).

Note that the minimum KL divergences for these corpora are larger than that of the fiction dataset, primarily because their texts are substantially shorter and therefore subject to stronger finite-size effects. In line with this, explicit random-tree simulations on the same texts yield finite-size baselines of \(0.133\) for \texttt{TinyStories}, \(0.181\) for \texttt{MCTest}, and \(0.152\) for \texttt{PoetryFoundationPoems}; the highlighted minima in Table~\ref{table:kl_div} are all comparable to these baselines.

The inferred \(K^\star\) values align with intuitive differences in textual complexity. Simpler corpora, such as AI-generated and human-written children's stories \cite{eldan2023tinystories,richardson2013mctest}, exhibit lower entropy rates and smaller \(K^\star\). Fiction \cite{fan2018hierarchical} occupies an intermediate regime, while modern poetry \cite{hf_suayptalha_poetry_foundation_poems} exhibits the largest entropy rates and largest effective branching factors. Thus, variation in entropy rate across genres is mirrored by variation in the structural complexity of their semantic trees.

\begin{figure*}
    \centering
        \includegraphics[width=0.85\textwidth]{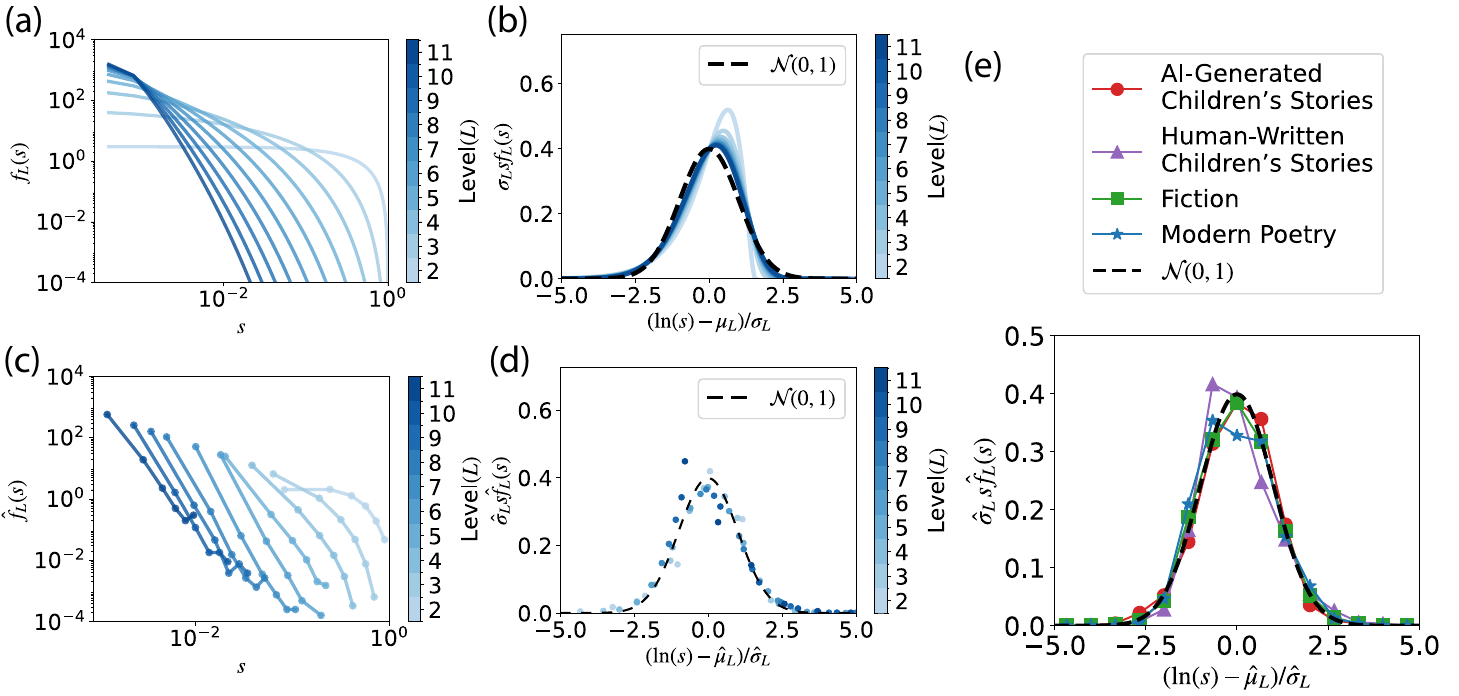}      
        \caption{\textbf{Semantic trees exhibit universal scaling across levels and corpora.} \\
        \textbf{(a)} Theoretical scaling functions $f_L$ across levels \(L\), shown on log--log axes.
        \textbf{(b)} When replotted using the \(O(1)\) lognormal variable \(x=(\ln s-\mu_L)/\sigma_L\), the curves in (a) collapse onto the universal \(\mathcal{N}(0,1)\) distribution as \(L\) increases. 
        \textbf{(c)} Empirical chunk size distributions \(\hat{f}_L\) (same as in Fig.~\ref{fig:tree_statistics}(b), rebinned to be uniform in \(\log s\)) across levels \(L\). 
        \textbf{(d)} The empirical curves in (c) likewise collapse under the transformation \(x=(\ln s-\hat{\mu}_L)/\hat{\sigma}_L\), consistent with the theoretical prediction. 
        \textbf{(e)} Empirical distributions from all four corpora collapse onto the same universal curve after standardization, suggesting that semantic trees share common scaling behavior despite differences in genre and effective branching factor. Each marker shows, for one corpus, the across-level mean histograms of the standardized variable as in (d), computed on a shared grid of 15 equal-width bins. 
        }
\label{fig:tree_universal}
\end{figure*}

\section{Scaling and universality of trees}
\label{sec:scaling}

In this section, we discuss a further theoretical consequence of the random tree model and test it against empirical semantic trees. After taking the large-\(N\) limit in Eq.~\eqref{eq:asymp_Pn}, the normalized chunk-size distribution depends only on \(L\) and \(K\) through the continuous scaling function \(f_L\). Using renormalization-group analysis (SI), we show that for \(L \gg 1\), \(f_L\) converges to a lognormal family whose parameters are determined by \(K\) and \(L\):
\begin{align}
    f_L(s) & \simeq \log \mathcal{N}(\mu_L, \sigma_L^2), \label{eq:lognormal}\\
    \mu_L &= -(L-1)H_{K-1},\\ 
    \sigma^2_L &= (L-1)H^{(2)}_{K-1},
\end{align}
where $H_{K-1}$ and $H^{(2)}_{K-1}$ denote the harmonic numbers of the first and second kind, respectively.

Equation~\eqref{eq:lognormal} implies a universality statement: after transforming to the \(O(1)\) lognormal variable
\begin{equation}
    x_L := \frac{\ln s_L - \mu_L}{\sigma_L},
\end{equation}
the level-dependent scaling functions \(f_L\) should collapse onto a universal standard normal distribution, \(\mathcal{N}(0,1)\), independent of \(L\) (and, after standardization, independent of \(K\) as well). Consistent with this prediction, the functions \(f_L\) differ substantially across levels when plotted in the original normalized-size coordinate \(s_L\) (Fig.~\ref{fig:tree_universal}(a)). However, when replotted as functions of \(x_L\), the same curves collapse onto \(\mathcal{N}(0,1)\) as \(L\) increases (Fig.~\ref{fig:tree_universal}(b)). We observe the same behavior empirically. The level-dependent empirical distributions in Fig.~\ref{fig:tree_universal}(c) collapse onto the same universal curve when expressed in the standardized coordinate \(x_L = (\ln s_L - \hat{\mu}_L)/\hat{\sigma}_L\), where \(\hat{\mu}_L\) and \(\hat{\sigma}_L\) are the empirical mean and standard deviation of \(\ln s_L\) at level \(L\) (Fig.~\ref{fig:tree_universal}(d)). For the empirical collapse, we standardize using the measured moments \(\hat{\mu}_L\) and \(\hat{\sigma}_L\), rather than the asymptotic expressions in Eq.~\eqref{eq:lognormal}. For finite texts, chunk sizes cannot fall below one token, imposing the cutoff \(\ln s_L \geq -\ln N\). Consequently, the level-wise mean and variance of \(\ln s_L\) saturate before reaching the asymptotic \(L\gg1\) regime predicted for an infinitely large tree. The universality prediction tested here concerns the shape of the distribution after standardization, rather than the finite-\(N\) values of its mean and variance.

Finally, the same standardized collapse is observed across corpora. When empirical chunk-size distributions from all four corpora are expressed in terms of the standardized log-size variable, they fall close to the same universal curve (Fig.~\ref{fig:tree_universal}(e)). Thus, although different corpora are characterized by different effective branching factors \(K^\star\) and different entropy rates, their semantic trees exhibit universal scaling behavior.

\section{Related work}
\label{sec:related_work}

The problem of representing linguistic structure has a long history in computational linguistics. Classical semantic parsing focuses primarily on sentence- and sub-sentence--level structure, where syntactic, logical, or grammatical constraints can often be evaluated against relatively well-defined formal criteria \cite{woods1973progress,zettlemoyer2012learning,mooney2007learning,li2020context}. Above the sentence level, discourse parsing provides a complementary view in which larger spans of text are organized by coherence relations, intentions, and rhetorical structure \cite{mann1988rhetorical,grosz-sidner-1986-attention,lascarides2007segmented}. However, such approaches often rely on expensive annotation and have historically been difficult to scale to large heterogeneous corpora \cite{carlsen2002rst,prasad2008pdtb2}.

The emergence of large language models has renewed interest in document-level segmentation, especially because many applications require processing texts that exceed a model's context window \cite{liu2024lost,lewis2020retrieval}. This has led to a family of methods commonly referred to as \emph{chunking}, which decompose long documents into smaller units for retrieval, summarization, or long-context reasoning \cite{wang2025document,hwang2025dynamic}. While fixed-size chunking partitions text into constant-length windows and may break semantic continuity, embedding-based and agentic semantic chunking methods attempt to identify contiguous, semantically coherent units that are more self-contained with respect to meaning and linguistic information \cite{qu2025semantic,kiss2025max,duarte2024lumberchunker}. Beyond flat chunking, hierarchical and tree-structured indexes have also been explored for retrieval and context assembly in retrieval-augmented generation \cite{sarthi2024raptor,rezazadeh2025isolated,wang2026alphacontext,zhao2026hierarchical,gupta2025llm,jin2025hierarchical,tao2025treerag}. Such hierarchical representations are also closely related to cognitive models describing how humans comprehend and recall narratives \cite{zhong2025random,georgiou2025large,thorndyke1975cognitive,baldassano2017discovering}. These developments show the practical value of hierarchical and semantic document structure, but they are not formulated as probabilistic ensembles from which an entropy rate can be computed.

A separate line of work has studied natural language through entropy, redundancy, surprisal, and perplexity. Since Shannon's original guessing-game estimates, entropy rate has served as a quantitative measure of linguistic uncertainty and predictability. Early reviews already emphasized that entropy rate is not a universal constant of a language, but varies with the statistical organization of the underlying material, including genre, style, domain, and text difficulty \cite{yaglom1983probability}. In particular, entropy and redundancy have been connected to graded reading materials, specialized languages, literary style, and other forms of textual organization. In psycholinguistics, per-token surprisal has been linked to behavioral measures of processing difficulty such as reading time and eye movements \cite{linzen2016uncertainty,smith2013logarithmic,demberg2008eyetracking,shain2024large}. More recently, neural language models have made it possible to estimate cross-entropy and perplexity across large corpora, renewing interest in how model surprisal reflects readability, corpus complexity, and systematic differences among text domains \cite{miaschi2020neural,miaschi2020perplexityreadability,wu2024perplexing,gao2020pile}. Recent work has also connected entropy rates to scaling laws in LLMs \cite{ho2024algorithmic,cagnetta2026deriving}.

However, the two perspectives discussed above has largely remained separate: Discourse parsing, semantic chunking, and hierarchical document representations describe how meaning is organized across scales, while entropy- and surprisal-based approaches quantify token-level predictability. What has been missing is a tractable framework that turns a semantic structure into a quantitative prediction of entropy rate. In this work, we address this gap by showing that the entropy from semantic trees contributes substantially to the overall entropy of language.

\section*{Discussion}

Our results connect two complementary views of natural language: language as a probabilistic token sequence and language as a hierarchy of coherent semantic units. In the first view, uncertainty is measured by token-level surprisal or cross-entropy under an auto-regressive model. In the second, meaning is organized across scales, with larger semantic units constraining the interpretation of smaller ones. The main contribution of this work is to show that these two views can be quantitatively linked: the multiscale semantic structure of a text contains enough information to predict a substantial component of its token-level entropy.

We established this link by recursively segmenting texts into coherent chunks, yielding semantic trees whose nodes represent text spans at multiple levels of granularity. We then modeled these trees with a one-parameter random \(K\)-ary tree ensemble. This ensemble is analytically tractable and predicts characteristic level-wise chunk-size distributions, which are observed in LLM-induced semantic trees across multiple open corpora. The entropy rate of the ensemble then provides an independent estimate from the LLM token-level surprisals.

Moreover, these two entropy estimates agree closely across corpora. Texts with lower LLM-measured entropy, such as children's stories, are also assigned lower structural entropy by the semantic tree ensemble; texts with higher LLM-measured entropy, such as poetry, correspondingly exhibit higher structural entropy. This suggests that corpus-level differences in entropy rate are not merely consequences of local lexical or syntactic statistics, but reflect differences in multiscale semantic organization. In this sense, entropy rate can be interpreted as the uncertainty over plausible continuations of a text's hierarchical semantic organization. 

The effective branching parameter \(K^\star\) also suggests a possible psycholinguistic interpretation. The random tree ensemble was first developed in the context of human narrative memory, where \(K\) plays the role of a working-memory capacity parameter \cite{zhong2025random}. In the present setting, \(K^\star\) may reflect working-memory load during comprehension, roughly, the number of concurrently active semantic chunks a reader must maintain to sustain coherent interpretation. Concretely, this could correspond to the number of keypoints needed to track a plot, or the number of interacting phrases required to resolve the meaning of a sentence. Notably, the empirically selected values \(K^\star \in [2,6]\) fall within the range typically associated with human working-memory limits \cite{cowan2010magical}. From this perspective, the perceived complexity of poetry relative to children's texts has a quantitative explanation: it induces a higher effective load on working memory during comprehension. Understanding the relationship between entropy rate, semantic complexity, and working memory remains an important direction for future work.

Several limitations should be emphasized. First, the LLM-based entropy rate is a model-dependent cross-entropy estimate, rather than a direct measurement of the true entropy rate of the underlying text distribution. Second, the semantic trees are themselves induced by LLMs, so their structure may reflect both properties of the texts and inductive biases of the chunking model. Third, assigning a single effective branching factor \(K^\star\) to an entire corpus is a coarse approximation: individual texts may vary across sections, levels, and genres in ways that are not captured by a one-parameter model. Accordingly, the agreement between LLM perplexity and tree-based entropy estimates should be interpreted primarily at the corpus level. Finally, the present theory captures the gross splitting behavior of the hierarchical chunk structure across scales, whereas the entropy rate can be sensitive to deviations in the splitting statistics at small chunk sizes. The theoretical consequences of such fine-scale deviations are analyzed in detail in \cite{sivan2026}; extending that analysis to empirical language data is left for future work. In addition, the theory does not explicitly model lexical choice, syntax, pragmatics, or world knowledge. Thus, the agreement between semantic tree entropy and LLM entropy should be interpreted as evidence that hierarchical semantic organization captures a major structural component of language uncertainty, rather than as a complete decomposition of all sources of linguistic entropy.

\vspace{+2em}
\begin{acknowledgments}
W.Z. acknowledges funding from NSF ACCESS Discover Grant CIS240836, and support from the Simons foundation and the Eric and Wendy Schmidt Membership in Biology at the Insitute for Advanced Study. MK is supported in part by a grant from Fran Morris Rosman and Richard Rosman. M.T. is supported by the Simons Foundation, MBZUAI-WIS Joint Program for Artificial Intelligence Research and Foundation Adelis. 
\end{acknowledgments}

\bibliographystyle{unsrt}
\bibliography{ref}

\clearpage

\pagebreak

\setcounter{page}{1}
\setcounter{equation}{0}
\setcounter{figure}{0}
\setcounter{section}{0}
\renewcommand{\theequation}{S.\arabic{equation}}
\renewcommand{\thefigure}{S\arabic{figure}}
\renewcommand*{\thepage}{S\arabic{page}}

\renewcommand{\sectionmark}[1]{}
\renewcommand{\subsectionmark}[1]{}

\onecolumngrid

\begin{center}
{\large \textbf{Supplementary Information for \\``\ourtitle"}}\\
\vspace{0.25cm}
\end{center}

\section{Markov chain with absorbing states}

Let's consider a tree model with following absorbing rules: (1) if a node fails to split at some level, i.e., all its sibling nodes are zero, or (2) if the node has size less than or equal to 1, then it stops splitting further (this is the model described in the PRL paper). 

To incorporate the absorbed states, for each node we can think of ``doubling" the original $N+1$ states with its accompanying absorbed states: the first $N+1$ states $0,...,N$ are the absorbed states, which does not split in the next level; $N+2, ..., 2N+2$ states are the survived states $0,...,N$, where we identify the fictitious survived $0,1$ states as identical to the absorbed $0,1$ states. We can define the state vector as the probability distribution of chunk sizes at the $l^{th}$ level, 
\begin{align}
\label{eq:state_vector}
    \vec{P}_L:= \left(P(n^{(l)}=0'),P(n^{(l)}=1'), \dots, P(n^{(l)}=N') | P(n^{(l)}=0),P(n^{(l)}=1), \dots, P(n^{(l)}=N) \right)^{\mathbf{T}}
\end{align}
where we use prime to denote the absorbed states. Then we can write down the Markov transition 
\begin{equation}
    \vec{P}_L = T^L \vec{P}_0
\end{equation} 
with the transition matrix given by

\begin{equation}\label{eq:T_absorb}
T \;=\;
\left(
\begin{array}{c|c}
\mathbf{1}_{N+1}
&
\begin{array}{ccccc}
1                      & 0                      & 0         & \cdots & 0                      \\
0                      & 1                      & 0         & \cdots & 0                      \\
0                      & 0                      & \dfrac1{Z_K(2)}  & \ddots & \vdots                 \\
\vdots                 & \vdots                 & \ddots    & \ddots & 0                      \\
0                      & 0                      & 0         & \cdots & \dfrac1{Z_K(N)}
\end{array}
\\\hline
\mathbf{0}_{N+1}
&
\begin{array}{ccccc}
0                                & 0                            & *                          & \cdots                    & *                       \\
0                                & 0                            & *                          & \dfrac{Z_{K-1}(j'-i')}{Z_K(i')} & \vdots                 \\
\vdots                           & \vdots                       & \ddots                     & \ddots                   & *                       \\
0                                & 0                            & \cdots                     & 0                         & 0
\end{array}
\end{array}
\right)
\end{equation}

where $i'=i-(N+1), j':= j-(N+1)$. The rows represent parent and columns represent their children, so all columns sum to one. The upper-left block simply means absorbed states stay absorbed. The lower-right block is our old upper-triangular Markov transition matrix with two new tweaks: (1) The first two columns are all zeros except for two entries (in the upper-right block) because we have identified the fictitious survived 0,1 states with the absorbed 0,1 states; (2) its diagonal are all zeros, because the diagonal entries represent the scenario where a node fails to split (size $n\to n$), therefore corresponds to a transition from survived state to absorbed state, and the previous diagonal entry $Z_K(i)^{-1}$ now simply gets moved to the diagonal in the upper-right block. 

\subsubsection{Occupation Number}
Let's call $T$ in Eq.~\eqref{eq:T_absorb} the node-size transition matrix. Instead of the node size distribution $\vec{P}_L$, we can also characterize the tree by its \textit{occupation number} $\vec{m}_L$, where $(m_L)_n:=$ number of nodes at level $L$ with size $n$. $\vec{m}_L$ is a $2(N+1)$ dimensional vector that characterizes the configurations of all the nodes at the same level $L$. For example, for $N=3$, $\vec{m}_L=(2,1,0,0,0,0,1,0)$ means that at level $L$, we have four nodes, three absorbed nodes with size 0, 0, 1, and one survived node with size 2. Note that the values sum up to 3. In general, normalization is enforced for the occupation number through
\begin{equation}
    N = \vec{\chi} \cdot \vec{m}_L, 
\end{equation}
where $\vec{\chi}=(0,1,...,N,0,1,...,N)$ is the node size vector. 

The occupation number $\vec{m}_L$ is also a Markov chain, 
\begin{equation}
    \vec{m}_L = Q^L \vec{m}_0,
\end{equation}
where $\vec{m}_0 = (0,...,0,1)$. The occupation transition matrix is related to the node-size transition matrix $T$ via a diagonal branching ratio matrix $B$, 
\begin{equation}
    Q = TB ; \;\; B_{ij}=\delta_{ij}b_j,
\end{equation}
where
\begin{align}
b_{i}= \begin{cases}
1, \quad i \in \text{absorbed}; \\
K, \quad \text{otherwise}.
\end{cases}
\end{align}

The total number of nodes up to and including level $L$ is given by
\begin{align}
\label{eq:C_absorb}
 C_L &= \vec{1} \cdot \vec{m}_L 
\end{align}
where $\vec{1}$ denotes a vector of $2(N+1)$ 1's (all the states in the system). 

\section{Large-$N$ limit}

For completeness, we reproduce the derivation first reported in \cite{zhong2025random} for taking the large-$N$ limit. 

In the limit of large $N$, and assuming $N\gg K$, we can use Stirling's approximation to write
\begin{equation}
\log Z_K(N) = \log \frac{N^{K-1}}{(K-1)!} + O\left(\log n\right).
\end{equation}
Then we can rewrite the conditional probability of observing $n^{(l+1)}$ stars in a bin of the $(l+1)^{th}$ level 
\begin{equation}
P(n^{(l+1)}|n^{(l)}) = \frac{Z_{K-1}(n^{(l)}-n^{(l+1)})}{Z_K(n^{(l)})}, 
\end{equation}
in terms of the normalized compression ratios $s_{l} = n^{(l)}/N$:
\begin{align}
    \rho(s_{l+1}| s_{l}) \equiv N P(N s^{(l+1)}| N s^{(l)}) =N \frac{Z_{K-1}(N (s_{l} - s_{l+1}))}{Z_{K}(N s_{l})} \approx (K-1)\frac{1}{s_{l}} \left( 1 - \frac{s_{l+1}}{s_{l}}\right)^{K-2}.
\end{align}
The chunk size distribution at level $L$ is given by marginalizing the intermediate chunks in the following product of conditional probabilities
\begin{align}
    P(n^{(L)}) &= \sum_{{\substack{1< n^{(L-1)} \\ < \dots < n^{(1)}}}}
    \prod_{l=1}^{L-1} P(n^{(l+1)}|n^{(l)})P(n^{(1)})  \label{appeq:marginal}\\
    &= \sum_{n^{(L-1)}} P(n^{(L)}|n^{(L-1)}) P(n^{(L-1)}), \label{appeq:markov_chain} \\
    P(n^{(1)}) &= \delta_{n^{(1)},N} \;.
\end{align}
The normalized chunk size distribution defined in Eq.~\eqref{eq:asymp_Pn} is given by
\begin{align}
\label{appeq:scaling_function_full}
    &f_L(s_L) = \prod_{l=1}^{L-1} \int_{s_{l+1}}^1 ds_l \rho(s_{l+1}|s_l) \rho(s_{1}|1) \\
    &\rho(s_{l+1}|s_l):= \frac{K-1}{s_l} \left(1-\frac{s_{l+1}}{s_l} \right)^{K-2}.
\end{align}

Therefore, the discrete distribution $P_L(n|N)$ converges to an \(N\)-independent continuous scaling function \(f_L(s)\), parameterized only by \(L\) and \(K\) \cite{zhong2025random}:

\begin{align}
    P_L(n|N) & \simeq \frac{1}{N} f_L(n/N), \label{eq:asymp_Pn1} \\
    f_{L}(s) &=
    \begin{cases}
        \displaystyle \int_s^1 \frac{dr}{r}\, p_B(r)\, f_{L-1}\!\left(\frac{s}{r}\right), & L>1,\\[2.2ex]
        \displaystyle \delta(s-1), & L=1.
    \end{cases}
\end{align}
Successive levels are related by a multiplicative convolution: the normalized child size is obtained by multiplying the parent size by a random fraction \(r\in(0,1)\), distributed as
\begin{align}
    p_B(r) = (K-1)(1-r)^{K-2},
\end{align}
corresponding to a \(\mathrm{Beta}(1,K-1)\) distribution.

\subsection{Scaling exponent for small $n$}
The scaling function $f_L(s_L)$ diverges as $s_L \to 0$, which allows us to simplify the nested integrals. 

As $s_L \to 0$, each of the integral $\int_{s_{l+1}}^1 ds_l/s_l(1-\frac{s_{l+1}}{s_l})^{K-2}$ is dominated by their lower limit, and becomes essentially $\int_{s_{l+1}}^1 ds_l/s_l$, therefore, we have very slow divergence

\begin{align}
    \lim_{s_L \to 0} f_L(s_L) =  \frac{(K-1)^{L-1}}{(L-2)!} \left[ \ln \frac{1}{s_L} \right]^{L-2}.
\end{align}

\subsection{Exact results for $K=2$}
\label{sec: K=2_exact}
The nested integral in Eq.~\eqref{appeq:scaling_function_full} can be written into a recursive form

\begin{align}
    f_{l}(s) &= \int_{s}^{1} \rho(s|s') f_{l-1}(s') ds'\\
    & = (k-1)\int_{s}^{1} dt \frac{(t - s)^{k-2}}{t^{k-1}} f_{l-1}(t)
\end{align}

For $l = 1$

\begin{align}
    f_{1}(s) = (k-1)(1 - s)^{k-2}
\end{align}

For $k = 2$, things simplify considerably.


\begin{align}
f_{1}(s) & = 1\\
f_{l}(s) & = \int_{s}^{1} dt  \frac{1}{t} f_{l-1}(t)
\end{align}

Seeing the first few cases instantly reveals the pattern

\begin{align}
    f_{2}(s) &= \int_{s}^{1} \frac{dt}{t} = - \log (s)\\
    f_{3}(s) & = \int_{s}^{1} \frac{dt}{t} \left( - \log (t)\right) =  \int_{s}^{1} dt \partial_{t} \left( - \frac{1}{2} \log^{2} t\right) = \frac{1}{2} \log^{2} t
\end{align}

Assume the ansatz
\begin{align}
    f_{l}(s) = c_{l} \left( - \log s\right)^{l-1}
\end{align}

Then
\begin{align}
    f_{l+1}(s) &=  c_{l}(-1)^{l-1} \int_{s}^{1} dt \partial_{t} \left(  \frac{1}{l} \log^{l} t\right)  = \frac{c_{l} (-1)^{l}}{l} \log^{l}(s) = c_{l+1} \left( - \log s\right)^{l}\\
    \Rightarrow & c_{l+1} = \frac{c_{l}}{l} = \frac{c_{l-1}}{l(l-1)}  = \frac{1}{\Gamma(l+1)}
\end{align}

So that $c_{l} = 1/\Gamma(l)$, and we arrive at the exact formula for $K=2$
\begin{align}
    f_{l}(s) = \frac{1}{\Gamma(l)} \left( - \log s\right)^{l-1}.
\end{align}

\section{Entropy of Trees}

For fixed \(K\), a tree realization can be represented by the collection of node sizes
\(T=(\mu_1,\dots,\mu_L) := (\mu_l)_{l=1}^{L}\), where \(\mu_{l;i}\) denotes the size of the \(i^{\mathrm{th}}\) node at level \(l\). Under the splitting process described above, the probability of a specific tree configuration factorizes as
\begin{align}
\label{eq:tree_prob}
    P(T) = \prod_{l=1}^{L} \prod_{i=1}^{K^{l-1}} Z_K(\mu_{l;i})^{-1}.
\end{align}
This form follows from the fact that, conditioned on a parent node size, the \(K\)-tuple of child sizes is uniformly distributed over weak ordered partition, and that there are \(K^{l-1}\) nodes at level \(l\) when zero-sized (empty) nodes are included. We emphasize, however, that conditioning on the root and marginalizing over intermediate levels induces highly non-uniform marginals such as \(P_L(n)\), as seen in main text Fig.~\ref{fig:tree_statistics}(a).

\subsection{Entropy decomposition}

Using Eq.~\eqref{eq:tree_prob}, we can write the Shannon entropy of the size-\(N\) random tree ensemble, i.e., the distribution over all tree configurations with root size \(N\), directly in terms of the level-wise chunk-size distributions \(P_L(n)\) (proof in Section~\ref{appsec:entropy_proof} below):
\begin{align}
    H(N) &= -\sum_{\substack{T \in \text{tree} \\\text{of size } N}} P(T) \log P(T) \\
        &= \sum_{L=1}^N \langle \log Z_K(n) \rangle_{n\sim \rho_L(n)}, \label{eq:entropy_decomp}
\end{align}
where $\rho_L(n)$ is the density of states measuring the number of internal nodes of size $n$ at level $L$,
\begin{align}
    \rho_L(n) &=
    \begin{cases}
        \displaystyle K^{L-1} P_L(n),& 2 \leq n \leq N-1 \\[2.2ex]
        \displaystyle 0,  & \text{otherwise}.
    \end{cases}
\end{align}
Equation~\eqref{eq:entropy_decomp} is exact and has an intuitive interpretation. By the Markov property of the splitting process, the tree entropy decomposes additively across levels and, within each level, into contributions from individual nodes. The per-node contribution, \(\log Z_K(n)\), is the log-multiplicity of possible \(K\)-way splitting configurations for an internal node of size \(n\). The factor \(\rho_L(n)\) weights this contribution by the number (multiplicity) of internal nodes of size \(n\) present at level \(L\).

\subsection{Proof of entropy decomposition (Eq.~\eqref{eq:entropy_decomp})}
\label{appsec:entropy_proof}
Given a tree configuration $(\mu_1, \dots, \mu_l)$, we simplify the notation by denoting $\mu_{l;i}$ as the size of the $i^{th}$ node at level $l$. The probability of $(\mu_1, \dots, \mu_l)$ is given by the product of the probabilities $q_{l-1}$ at each \textit{preceding} level $l-1$,
\begin{align}
    P(\mu_1, \dots, \mu_l) :&= \prod_{m=0}^{l-1} q_m; \\
    q_m := \frac{1}{Z_K(\mu_m)}&, \;\;\; q_0 = \frac{1}{Z_K(N)}.
\end{align}
At each level $l+1$, $q_l$ describes the uniform probability across all the children nodes splitted from their parents at level $l$. Therefore, $q_l$ is given by the product of the probabilities across all the nodes at level $l$
\begin{align}
\label{eq:prob_level}
    q_l = \prod_{i=1}^{K^l} \frac{1}{Z_K(\mu_{l;i})}.
\end{align}
The Shannon entropy at level $l+1$ is given by 
\begin{align}
    H_{l+1} &= -\sum_{\mu_1 \dots \mu_{l+1}} P(\mu_1, \dots, \mu_{l+1}) \log P(\mu_1, \dots, \mu_{l+1}) \\
    &= -\sum_{\mu_1 \dots \mu_{l+1}} q_0 \cdots q_{l} \log (q_0 \cdots q_{l}) \\
    &= -\sum_{\mu_1 \dots \mu_{l}} \left(q_0 \cdots q_{l-1}\right) \left(\sum_{\mu_{l+1}}q_l \right) \log \bigg( (q_0 \cdots q_{l-1}) q_l \bigg)
\end{align}
where we note that the first parenthesis is $P(\mu_1, \dots, \mu_{l})$ and second parenthesis equals one by definition of $q_l$. We can rewrite
\begin{align}
\label{eq:entropy_increase}
    H_{l+1}&= -\sum_{\mu_1 \dots \mu_{l}} P(\mu_1, \dots, \mu_{l})\log P(\mu_1, \dots, \mu_{l}) - \sum_{\mu_1 \dots \mu_{l}}P(\mu_1, \dots, \mu_{l})\log q_l \\
    &= H_l + \Delta H_{l+1}.
\end{align}
Using Eq.~\eqref{eq:prob_level},
\begin{align}
    \Delta H_{l+1} &= - \sum_{\mu_1 \dots \mu_{l}}P(\mu_1, \dots, \mu_{l})\log q_l \\
    &= \sum_{\mu_1 \dots \mu_{l}}P(\mu_1, \dots, \mu_{l}) \sum_{i=1}^{K^l}\log Z_K(\mu_{l;i}),
\end{align}
For reasons that will become clear later, let's change the name of $\mu_{l;i}$ by introducing a new variable $n_l$ using $\sum_{n_l} \delta({n_l - \mu_{l;i}})$. Then
\begin{align}
\label{eq:deltaH_no_absorb}
    \Delta H_{l+1} &= \sum_{n_l} \sum_{\mu_1 \dots \mu_{l}}P(\mu_1, \dots, \mu_{l}) \sum_{i=1}^{K^l}\log Z_K(n_l) \delta({n_l - \mu_{l;i}}) \\
    &= K^l \sum_{n_l} \bigg[\sum_{\mu_1 \dots \mu_{l}}P(\mu_1, \dots, \mu_{l}) \bigg(\frac{1}{K^l} \sum_{i=1}^{K^l}\delta({n_l - \mu_{l;i}}) \bigg) \bigg] \log Z_K(n_l) \\
    &= K^l \sum_{n_l} \left\langle \frac{1}{K^l} \sum_{i=1}^{K^l}\delta({n_l - \mu_{l;i}}) \right\rangle_{\substack{ P(\mu_1, \dots, \mu_{l})}} \log Z_K(n_l) \\
    &= K^l \sum_{n_l} \left\langle \langle \delta(n_l - \mu_l) \rangle_{\substack{\text{bins at}\\ \text{level } l}} \right\rangle_{\substack{\text{trees up}\\ \text{to level } l}} \log Z_K(n_l) \\
    &= K^l \sum_{n_l} P_l (n_l) \log Z_K(n_l) \\
    &= K^l \sum_n \langle \log Z_K(n) \rangle_{P_l(n)}.
\end{align}
where we have used the definition of the marginal distributions of node size at level $l$.

Since entropy comes from just the survived states, we define as in Eq.~\eqref{eq:entropy_decomp} $\rho_L(n):=K^L P_L(n)$ for $2 \leq n \leq N-1$, we can simply write
\begin{equation}
\label{eq:entroy_dbl_sum}
    H(N) = \sum_{L=1}^{N} K^L \sum_{n = 2}^{N-1} P_L(n) \log Z_K(n).
\end{equation}
Here we enumerate all the trees for a few cases by hand, and computed their entropy, which is reported below (the notation $(n_1,...,n_m)^r$ denotes the partition $(n_1,...,n_m)$ with $r$ permutations):
\begin{itemize}
    \item $N=3$, $K=2$. A total of 8 trees: $(3,0)^2$ with $p=1/4$; $\big( (2,0)^2, 1\big)^2$ with $p=1/12$; $\big( (1,1), 1\big)^2$ with $p=1/12$. Entropy is $2\log2+1/2\log3 \approx 1.936 \text{ nats}$. 
    \item $N=4$, $K=3$. A total of 369 trees. Entropy is 5.134\text{ nats}. 
    \item $N=5, K=3$. A total of 3885 trees. Entropy is 6.983\text{ nats}.
\end{itemize}

\begin{figure*}[ht]
    \centering
    \includegraphics[width=0.35\textwidth]{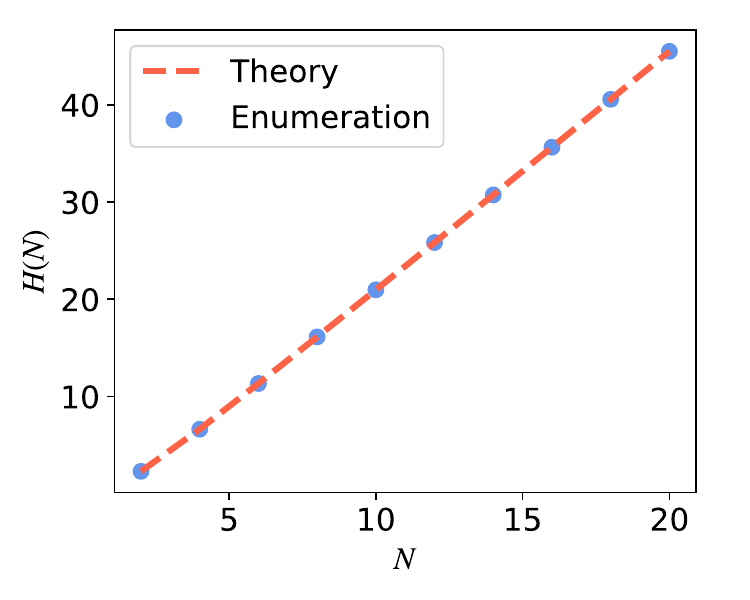}
    \caption{\textbf{Entropy of trees.} Theory and enumeration agrees for $K=4$, and gives $H(N) \approx 2.5 \text{ nats} \times N$. }
    \vspace{-1em}
\label{fig:entropy_enumeration}
\end{figure*}

These all agree with the numerical values computed from Eq.~\eqref{eq:entroy_dbl_sum}. We also enumerate the trees using a program and compare the entropy with Eq.~\eqref{eq:entroy_dbl_sum}. The result agrees exactly as shown in Fig.~\ref{fig:entropy_enumeration}. The curve is computed for $K=4$, it has some initial nonlinearity (for $N<5$) but soon becomes linear, with slope about 2.5 $\text{nats}$. 

\subsection{Entropy recursion}

We can rewrite the entropy decomposition formula Eq.~\eqref{eq:entroy_dbl_sum} into a recursive one (as in main text Eq.~\eqref{eq:H_LHS_main}), which is useful for calculating the entropy numerically. 
\begin{align}
\label{eq:H_LHS}
    H(N) = \ln Z_K(N) + K \sum_{n=2}^{N-1} \frac{Z_{K-1}(N-n)}{Z_K(N)} H(n).
\end{align}
To see Eq.~\eqref{eq:H_LHS} and Eq.~\eqref{eq:entroy_dbl_sum} are equivalent, we repeatedly substitute the $H(N)$ from the RHS of Eq.~\eqref{eq:H_LHS} into the LHS for $N-1$ times and organize terms in powers of $K$:
\begin{align}
    H(N) &= K^0 \ln Z_K(N) \\
    &+ K^1 \sum_{n_1=2}^{N-1}\frac{Z_{K-1}(N-n_1)}{Z_K(N)}\ln Z_K(n_1) \\
    &+ \cdots \\
    &+ K^{N-1} \sum_{n_1=2}^{N-1} \cdots \sum_{n_{N-1}=2}^{n_{N-2}-1} \left[ \frac{Z_{K-1}(N-n_1)}{Z_K(N)} \cdots \frac{Z_{K-1}(n_{N-2}-n_{N-1})}{Z_K(n_{N-2})}\right] \ln Z_K(n_{N-1})  \\
    &+ K^N \sum_{n_1=2}^{N-1} \cdots \sum_{n_N=2}^{n_{N-1}-1} \left[ \frac{Z_{K-1}(N-n_1)}{Z_K(N)} \cdots \frac{Z_{K-1}(n_{N-1}-n_N)}{Z_K(n_{N-1})}\right] H(n_N).
\end{align}
However, note that $H(n_N)=0$, because each survived children chunk $n_D$ is at least size 1 smaller than its parent chunk, $n_{D-1} - n_D \geq 1$, otherwise $n_D$ will be an absorbed state. So $n_1 - n_N \geq N-1$, and $n_N \leq 1$ is an absorbing state. We recognize the factions are conditional probabilities 
\begin{align}
\label{eq:conditional}
    P(n_D|n_{D-1}) &= \frac{Z_{K-1}(n_{D-1}-n_{D})}{Z_K(n_{D-1})},
\end{align}
therefore for the $O(K^{D})$ term we have
\begin{align}
    &\sum_{n_1=2}^{N-1} \cdots \sum_{n_{D}=2}^{n_{D-1}-1} \left[ \frac{Z_{K-1}(N-n_1)}{Z_K(N)} \cdots \frac{Z_{K-1}(n_{D-1}-n_{D})}{Z_K(n_{D})}\right] \ln Z_K(n_{D}) \\
    &= \sum_{\substack{2 \leq n_{D} \leq n_{D-1}\\
    < \cdots < n_1 \leq N-1 }} P(n_1|N) \cdots P(n_{D}|n_{D-1}) \ln Z_K(n_{D}) \\
    &= \sum_{n_{D} =2}^{N-1} P_{D}(n_{D}|N) \ln Z_K(n_{D}),
\end{align}
where in the last equality we have used the law of total probability (also see Eq.~\eqref{appeq:markov_chain}). 

\subsection{Entropy scaling}
Below, in Sec.~\ref{sec:K2} we first demonstrate a simple derivation of the entropy rate for the case of $K=2$ as an illustration. In Sec.~\ref{sec:largeK_limit}, we show that in the limit of $K\gg 1$ it is possible to prove that $H(N)$ scales linearly in $N$ (Eq.~\eqref{eq:entropy_extensive}) and obtain close-form solution for $h_K$.

\subsubsection{$K=2$ (exact)}
\label{sec:K2}
For the specific case of $K=2$, the partition functions simplify to $Z_2(N) = N+1$ and $Z_1(N-n)=1$. Substituting these into the general recursion Eq.~\eqref{eq:H_LHS}, we obtain:
\begin{align}
    H(N) = \ln(N+1) + \frac{2}{N+1} \sum_{n=2}^{N-1} H(n).
    \label{eq:H_recurrence_K2}
\end{align}

This recurrence relation has the same form as the one governing the average number of comparisons in the ``quicksort'' array sorting algorithm, which can be solved exactly using a differencing trick, as shown by Knuth \cite{knuth1998art}. Cover and Thomas were the first to apply the same technique to calculate the entropy of random binary trees, given as an exercise in their seminal book on information theory (Problem 4.5 in \cite{cover1999elements}). While their exercise corresponds to random strong integer ordered partition with $K=2$, here we apply it to our weak ordered partition random tree model with $K=2$. Despite this difference in the underlying combinatorics, the form of the recurrence remains the same, and thus can be solved using the same differencing trick.

The first step is to get rid of the summation term by considering the difference $(N+1)H(N) - N H(N-1)$, which leads to an exactly solvable first-order linear recurrence relation. The solution is given by:
\begin{align}
    H(N) = \ln(N+1) + 2(N+2) \sum_{m=2}^{N-1} \frac{\ln(m+1)}{(m+2)(m+3)}.
\end{align}

In the limit of large $N$, the entropy scales linearly as $H(N) \sim h_2 N$. Extracting the coefficient of the linear term, we find the exact entropy rate for $K=2$:
\begin{align}
    h_2 = 2 \sum_{m=2}^{\infty} \frac{\ln(m+1)}{(m+2)(m+3)}.
\end{align}

\subsubsection{$K \gg 1$}
\label{sec:largeK_limit}

Recall that the normalized chunk size distribution can be written as a Mellin convolution with beta-variables, Eq.~\eqref{eq:Mellin_conv}, we can formally solve it by first applying the Mellin transform, then perform the inverse Mellin transform using contour integral. However, as will be noted at the end of this section, this argument only works for $K \gg 1$. 

For a single Beta-variable $r \sim p_B(r)$, the Mellin transform of its pdf is
\begin{align}
    \mathcal{M}[p_B](z) &= \int r^{z-1}p_B(r)dr \\
    &= (K-1)B(z,K-1),
\end{align}
where $B$ is the Euler Beta function. Our distribution $f_L(s)$ describes the product of $L-1$ i.i.d. random beta variables (Eq.~\eqref{eq:s_product}), therefore the pdf in the Mellin space multiplies, and we have
\begin{align}
\label{eq:forward_mellin}
    \mathcal{M}[f_L](z) &= \left(\mathcal{M}[p_B]\right)^{L-1} \\
    &= (K-1)^{L-1}B(z,K-1)^{L-1},
\end{align}
which also means the moments of $f_L$ are (by definition of Mellin transform)
\begin{align}
    m_t = (K-1)^{L-1}B(t+1,K-1)^{L-1}.
\end{align}
Applying the inverse Mellin transform to Eq.~\eqref{eq:forward_mellin}, we have the spectral representation of the normalized chunk size distribution 
\begin{align}
\label{eq:spectral_f}
    f_L(s) = \frac{1}{2\pi i}\int_{c-i\infty}^{c+i\infty} dz s^{-z} (K-1)^{L-1}B(z,K-1)^{L-1},
\end{align}
where $c$ is taken to be some positive constant. 

Note that when taking the large $N$ limit, the chunk size distribution $P_L(n)$, which is a discrete probability mass function, cannot converge point-wise to the p.d.f. $f_L(s)$ because the support is different. To enforce normalization we assume the bin-wise convergence:
\begin{align}
    P_L(n) &= \int^n_{n-1} \frac{1}{N}f_L\left(\frac{n'}{N}\right)dn' \\
    &= \int^{n/N}_{(n-1)/N} f_L(s) ds.
\end{align}
Then in the large-$N$ limit, the entropy Eq.~\eqref{eq:entroy_dbl_sum} becomes
\begin{align}
    H(N) &= \frac{1}{N}\sum_{n =2}^{N-1} \ln Z_K(n) g\left(\frac{n}{N} \right), \label{eq:S_gn} \\
    g\left(\frac{n}{N}\right) &= N\sum_{L=1}^N K^{L-1} \int^{n/N}_{(n-1)/N} f_L(s) ds. \label{eq:gnN}
\end{align}
Using Eq.~\eqref{eq:spectral_f}, Eq.~\eqref{eq:gnN} becomes
\begin{align}
    g\left(\frac{n}{N}\right) &= \frac{1}{2\pi i} \int^{n/N}_{(n-1)/N} ds\int_{c-i\infty}^{c+i\infty} dz s^{-z} \sum_{L=1}^N \left[K(K-1)B(z,K-1) \right]^{L-1} \\
    &= \frac{1}{2\pi i} \int^{n/N}_{(n-1)/N} ds \int_{c-i\infty}^{c+i\infty} dz \frac{s^{-z}}{1 - K(K-1)B(z, K-1)}. \label{eq:contour}
\end{align}
The integrand
\begin{align}
    F(z):=\frac{s^{-z}}{1 - K(K-1)B(z, K-1)}
\end{align}
takes the form 
\begin{align}
    F(z) &= \frac{s^{-z}}{h(z)}; \\
    h(z)&:= 1 - K(K-1)B(z, K-1).
\end{align}
The poles in $F$ corresponds to the zeros $z_j \in \mathbb{R}$ of $h(z)$
\begin{align}
    1 = K(K-1)B(z_j, K-1),
\end{align}
which exists in the complex plane leftward of $z=2$, and $\text{max}_j \text{Re}(z_j) = 2$. We can perform a Laurent expansion of the denominator $h(z)$ around the zeros
\begin{align}
    h(z) &= h'(z_j)(z-z_j) + O((z-z_j)^2); \\
    h'(z) &= -K(K-1)B(z,K-1)\left[\psi(z) - \psi(z+K-1) \right], \label{eq:hz}
\end{align}
where $\psi(x):= d\ln \Gamma(x)/dx$ is the digamma function. In particular, at the zeros $h(z_j)=0$, the derivative simplifies to
\begin{align}
    h'(z_j) = \psi(z_j + K-1) - \psi(z_j).
\end{align}
For integer $n_j$, it further simplifies into Harmonic numbers
\begin{align}
    h'(n_j) &= H_{n_j + K -2} - H_{n_j-1}, \\
    H_K &= \sum_{i=1}^K \frac{1}{i}.
\end{align}
All the poles are toward the left of $z=2$, some are complex some are real, depending on choice of $K$. Eq.~\eqref{eq:contour} is a Bromwich type integral, and we choose $c>2$ and close the contour counterclockwise to the left with radius $R\to \infty$, where the integrand decays exponentially like $e^{\ln s R}$ ($\ln s\leq 0$). The contour therefore encloses all the poles of $F(z)$. By the Residue Theorem, we can write the integral Eq.~\eqref{eq:contour} as
\begin{align}
    g\left(\frac{n}{N}\right) &= \int^{n/N}_{(n-1)/N} ds\sum_k \text{Res}(F, z_k). \label{eq:g_res}
\end{align}
The residue at the dominant pole ($z=2$) becomes
\begin{align}
    &\int^{n/N}_{(n-1)/N} ds \text{Res}(F,z=2) \\
    = &\frac{N}{H_K - 1}  \int^{n/N}_{(n-1)/N} s^{-2} ds \\
    = &\frac{N^2}{(H_K - 1)}\frac{1}{n(n-1)}.
\end{align}
All other poles are sub-leading in $1/N$, $\text{Re}(z_j) < 2$, and their contributions $s^{-z_j} = (N/n)^{z_j}$ are exponentially smaller in $N$. All the poles are simple poles, because $h(z_j) = 0$ but $h'(z_j)\neq 0$. We can in fact calculate them in closed form. Let's rewrite Eq.~\eqref{eq:hz} using Gamma function recurrence relations
\begin{align}
    h(z) &= 1 - \frac{K!}{Q(z)}, \\
    h'(z) &= K! \frac{\sum_{m=0}^{K-2} (z+m)^{-1}}{Q(z)} \\
    Q(z) &= \prod_{m=0}^{K-2}(z+m).
\end{align}
Then the poles are simply the $K-1$ complex roots of the polynomial equation
\begin{align}
    K! = Q(z).
\end{align}
At these roots, the derivative simplifies to 
\begin{align}
    h'(z_j) = \sum_{m=0}^{K-2} \frac{1}{z_j + m}.
\end{align}
Therefore, the residues are 
\begin{align}
    \text{Res}(F,z_j) = \frac{s^{-z_j}}{\sum_{m=0}^{K-2}(z_j+m)^{-1}}.
\end{align}
Then we have the asymptotic form of entropy
\begin{align}
    H(N) &= \frac{1}{N}\sum_{n=2}^{N-1} \ln Z_K(n) g\left(\frac{n}{N}\right),  \\
    g\left(\frac{n}{N}\right) &= \int^{n/N}_{(n-1)/N} ds \sum_{j=0}^{K-1} \frac{s^{-z_j}}{\sum_{m=0}^{K-2}(z_j+m)^{-1}}, \\
    z_j  &\in \text{ roots of } 0 = \prod_{m=0}^{K-2}(z+m) - K!.
\end{align}
For $N \gg 1$, the dominant contribution is $z=2$. Going back to Eq.~\eqref{eq:S_gn} we have
\begin{align}
\label{eq:ck_before}
    H(N) &= h_K N + O(\log N), \\
    h_K &= \frac{1}{H_K - 1}\sum_{n=2}^{N-1} \frac{\ln Z_K(n)}{n(n-1)}. \label{eq:ck_after}
\end{align}

Recall that when deriving the recurrence relation for the pdf $f_L(s)$ in Eq.~\eqref{eq:beta_kernel}, we used Stirling approximation for $n \gg1 $ (Eq.~\eqref{eq:conditional}):
\begin{align}
    Z_K(n)&=\binom{n+K-1}{K-1} \\
    &\approx \frac{n^{K-1}}{(K-1)!}.
\end{align}
This approximation could fail in $Z_{K-1}(n_{L-1} - n_L)$ when $n_{L-1} \sim n_L$. However, it becomes accurate when $K\gg1$, because the typical size of $n_L$ is 
\begin{align}
    \langle n_L \rangle \sim \frac{n_{L-1}}{K}.
\end{align}
Therefore, for large $K$, $n_L \ll n_{L-1}$ for most of the configurations. For small $K$, however, the edge contribution $n_{L-1} \sim n_L$ become non-negligible, and the error builds up. Away from the edge, the error is 
\begin{align}
    P(n|N) = \frac{K-1}{N} \left(1-\frac{n}{N}\right)^{K-2} + O\left(\frac{1}{N^2}\right).
\end{align}
and subsequently
\begin{align}
    P_L(n) = \int^{n/N}_{(n-1)/N} f_L(s)ds + O\left(\frac{L}{N^2}\right),
\end{align}
which carries to the entropy 
\begin{align}
    H(N) = h_K N + O(\log N).
\end{align}

\begin{figure*}[ht]
    \centering
    \includegraphics[width=0.35\textwidth]{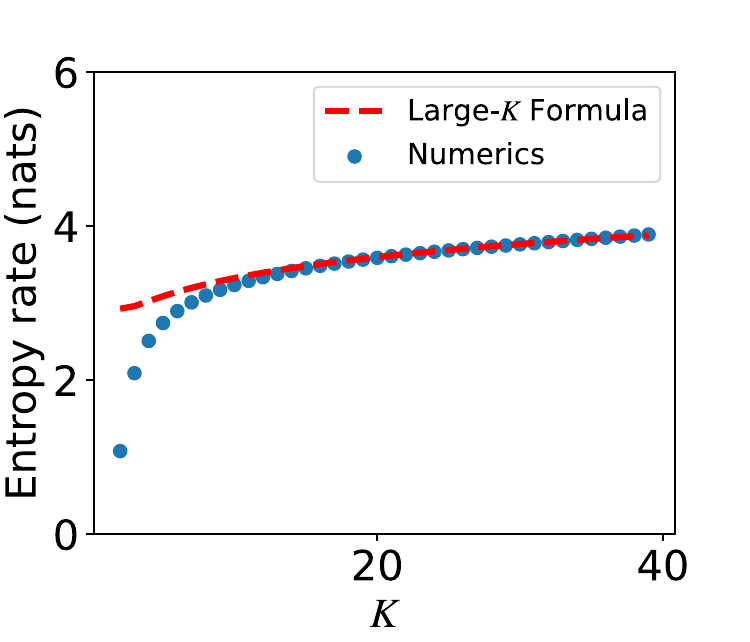}
    \caption{\textbf{Entropy scaling}. Numerics is computed with the Markov model, large $K$ expansion is Eq.~\eqref{eq:c_largeK}. }
    \vspace{-1em}
\label{fig:entropy_scaling}
\end{figure*}

To see how good the approximation is, we expand Eq.~\eqref{eq:ck_after} for large $K$ ($\gamma \approx 0.577$ is Euler's constant)
\begin{align}
\label{eq:c_largeK}
    h_K = \frac{1}{2} (\ln K)^2 + (1+\gamma)\ln K + \frac{\pi^2}{12} - \ln 2 + O(\frac{\ln K}{K}).
\end{align}
In Fig.~\ref{fig:entropy_scaling} we plot the numerics against Eq.~\eqref{eq:c_largeK}, it appears to be an good approximation for $K \gg 1$.

\section{Recursive semantic chunking}

\subsection{Overview}
Before chunking, we sanitize the user text by stripping an outer triple-backtick fence (if present), normalizing line endings, and converting tabs/newlines to single spaces. We then enforce sentence spacing by inserting a space after ".", "!" or "?" when the next character appears to start a new sentence (capital letter or digit), collapse multiple spaces into one, and trim leading/trailing whitespace. Finally, we remove control characters and delete all single/double quotation marks (ASCII and common Unicode variants), producing a stable, quote-free string whose remaining punctuation and interior spacing are consistent for both the chunker and verifier. 

Then the text is passed into the chunking prompt (see below), and input to a LLM (\texttt{LLama-4-Maverick} or \texttt{DeepSeek-V3.1}) for segmentation. After receiving results returned from the chunker, they are pass into a verifier, where the verifier concatenate the chunks and check whether it equals the original text before chunking. 

\subsection{Chunking routines}

We use three chunking modes to balance minimal assumptions, robustness, and token efficiency across regimes. For most inputs, we default to sentence-level segmentation because it makes the weakest structural assumptions while still producing semantically coherent spans. For long texts, however, LLMs occasionally skip or paraphrase content, which breaks exact-match verification; to reduce this failure mode and to save input budget, we switch to a paragraph-level cut-point mode where the model only returns sentence cut point indices, and we only trigger this once the span exceeds a 200-token threshold, which is well above single-sentence level, so we are not imposing fine-grained sentence structure at that level. Finally, for very short spans (fewer than 6 tokens), models often resist splitting within words and may stop before reaching token granularity; in these cases, we provide an explicit token-indexed list and use a phrase-level cut-point prompt so the model can operate directly on token boundaries.

Below, we show the three different chunking prompts we use at the different levels:

\begin{prompt}[prompt:segmentation]{Text Segmentation (main)}
\subsubsection*{System Prompt}

\texttt{You are a text-segmentation assistant. 
Split the entire input into up to *\{K\}* contiguous, non-overlapping segments such that concatenating them in order reproduces the original text exactly. Including ALL whitespaces and punctuation.
Each segment should be semantically coherent.
In the case of simple phrases, segment into semantically contiguous chunks of tokens.
*CRITICAL*: Preserve ALL whitespace and quotation marks exactly as it appears in the input. If the input starts with a space, the first segment should start with that space.
Return a raw JSON array of strings - do NOT wrap in markdown or code fences.}
\\
\subsubsection*{Examples}

\noindent\textbf{Input:}
\begin{lstlisting}[language=json,firstnumber=1]
"The quick brown fox jumps over the lazy dog."
\end{lstlisting}

\noindent\textbf{Output:}
\begin{lstlisting}[language=json,firstnumber=1]
["The quick brown fox", " jumps over", " the lazy dog."]
\end{lstlisting}

\noindent\textbf{Input:}
\begin{lstlisting}[language=json,firstnumber=1]
" brown fox"
\end{lstlisting}

\noindent\textbf{Output:}
\begin{lstlisting}[language=json,firstnumber=1]
[" brown", " fox"]
\end{lstlisting}

\noindent\textbf{Input:}
\begin{lstlisting}[language=json,firstnumber=1]
"the lazy dog."
\end{lstlisting}

\noindent\textbf{Output:}
\begin{lstlisting}[language=json,firstnumber=1]
["the", " lazy dog", "."]
\end{lstlisting}

\noindent\textbf{Input:}
\begin{lstlisting}[language=json,firstnumber=1]
" jumps over"
\end{lstlisting}

\noindent\textbf{Output:}
\begin{lstlisting}[language=json,firstnumber=1]
[" jumps", " over"]
\end{lstlisting}

\subsubsection*{User Prompt}
Input text to segment:

\begin{lstlisting}[language=json]
{text}
\end{lstlisting}

\end{prompt}

\begin{prompt}[prompt:cut_points]{Paragraph Segmentation (Cut-Point Mode)}
\subsubsection*{System Prompt}

\texttt{You are a text-segmentation assistant **in CUT-POINT mode**.
The input will be N numbered sentences (0 \ldots\ N-1).
Choose up to *\{K-1\}* cut points so that the resulting segments
are semantically contiguous and most conceptually disjoint from the other segments.
Return **raw JSON** -- a strictly ascending list of integers in the
range 1 \ldots\ N-1. Do NOT wrap in markdown.}
\\
\subsubsection*{Example}

\noindent\textbf{Input:}
\begin{lstlisting}[language=json,firstnumber=1]
[0] Yeah I was in the boy scouts at the time.
[1] And we was doing the 50-yard dash racing but we was at the pier marked off and so we was doing the 50-yard dash.
[2] There was about 8 or 9 of us you know, going down, coming back.
[3] And going down the third time I caught cramps and I started yelling 'Help!' but the fellows didn't believe me you know.
[4] They thought I was just trying to catch up because I was going on or slowing down.
\end{lstlisting}

\noindent\textbf{Output:}
\begin{lstlisting}[language=json,firstnumber=1]
[1, 3]
\end{lstlisting}

\subsubsection*{User Prompt}
Input numbered sentences:

\begin{lstlisting}[language=json]
{sentences}
\end{lstlisting}

\end{prompt}
\begin{prompt}[prompt:phrase_cut_points]{Phrase-Level Segmentation (Token Indices)}
\subsubsection*{System Prompt}

\texttt{You are a **phrase-level segmentation assistant**.
Given a token list with indices, return a JSON array of cut-points (0-based) where a new segment starts.
Choose no more than *\{K-1\}*, yielding up to *\{K\}* segments in total.
Segments should be the largest contiguous group of tokens that form a coherent phrase (e.g. determiner + noun, adjective + noun, verb + particle).
Attach sentence-final punctuation to the preceding word.
Cut-points must be strictly increasing and between 1 and len(tokens)-1.
**JSON only** -- respond with the RAW JSON array (e.g. [3]) and nothing else.}
\\
\subsubsection*{Example}

\noindent\textbf{Input:}
\begin{lstlisting}[language=json,firstnumber=1]
{"tokens": ["The", " quick", " brown", " fox"]}
\end{lstlisting}

\noindent\textbf{Output:}
\begin{lstlisting}[language=json,firstnumber=1]
[3]
\end{lstlisting}

\noindent\textbf{Implied Segments (for illustration):}
\begin{lstlisting}[language=json,firstnumber=1]
["The quick brown", " fox"]
\end{lstlisting}

\subsubsection*{User Prompt}
Input token list:

\begin{lstlisting}[language=json]
{tokens}
\end{lstlisting}

\end{prompt}

\begin{figure*}
    \centering
    \includegraphics[width=1\textwidth]{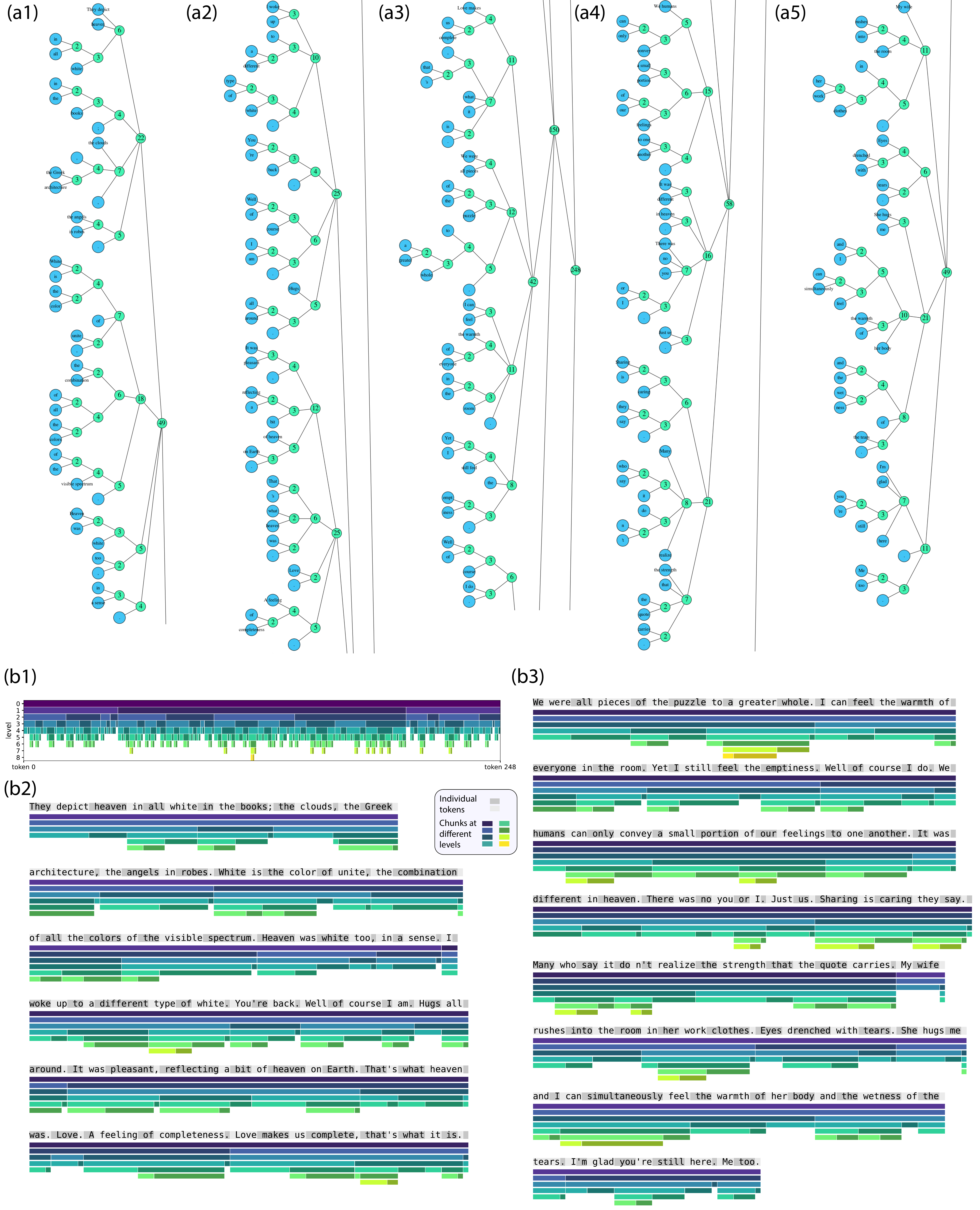}
    \caption{\textbf{Semantic tree of \texttt{RedditStories} 32721.} The tree was originally plotted as a long vertical figure and is adapted here into five panels. Read the tree from top to bottom and left to right across panels (a)--(e). The root appears in panel (c) as the green node (248).}
    \vspace{-1em}
\label{fig:example_reddit_story}
\end{figure*}

\subsection{Chunk-size statistics}

The mean KL divergence $\langle D_{\text{KL}} \rangle$ reported in Table~\ref{table:kl_div} is computed from 50 stories per corpus. Because the depth of a semantic tree depends on both the text and the chunking hyperparameter $K$, no single cutoff level is equally well sampled across conditions: small-$K$ trees branch more slowly and therefore run deeper, so at any fixed level they retain more chunks than large-$K$ trees, whose deep levels are supported by only a handful of surviving nodes. Averaging up to a universal cutoff would then favour small $K$, whose per-level estimates remain well resolved while those at large $K$ are dominated by finite-sample noise. We therefore threshold on statistics rather than on depth, averaging over every level at or beyond level 2 (the first non-trivial level, level 1 is the entire text) that contains at least as many chunk-size samples as level 2.

\subsection{Examples semantic trees}

In this section, we give a few examples of the resulting semantic trees from the algorithm described above. 

\subsubsection*{Example \texttt{RedditStories} (32721)}
Due to the space constraints, we present a shorter story below (248 tokens), while a typical story are much longer. The following is an example (32721) completed by a Reddit user in response to another user-written prompt. The whitespaces and linebreaks have been normalized for display in below. The corresponding semantic tree for $K=4$ is shown in Fig.~\ref{fig:example_reddit_story}.

\subsubsection*{Prompt}
\begin{quote}
``You have been enjoying Life in Heaven for over 100 Years and then suddenly, you wake up on a hospital bed \ldots''
\end{quote}

\subsubsection*{Story Excerpt}
\begin{quote}
They depict heaven in all white in the books; the clouds, the Greek architecture, the angels in robes. White is the
color of unity, the combination of all the colors of the visible spectrum. Heaven was white too, in a sense.

I woke up to a different type of white.

You're back. Well of course I am. Hugs all around. It was pleasant, reflecting a bit of heaven on Earth. That's what
heaven was. Love. A feeling of completeness. Love makes us complete, that's what it is. We were all pieces of the puzzle
to a greater whole.

I can feel the warmth of everyone in the room. Yet I still feel the emptiness. Well of course I do. We humans can only
convey a small portion of our feelings to one another. It was different in heaven. There was no you or I. Just us.
Sharing is caring they say. Many who say it don't realize the strength that the quote carries.

My wife rushes into the room in her work clothes. Eyes drenched with tears. She hugs me and I can simultaneously feel
the warmth of her body and the wetness of the tears.

I'm glad you're still here.
\end{quote}

\subsubsection*{Example \texttt{TinyStories} (198810)}

In the following we give an example from the \texttt{TinyStories} (198810), it has a total of 135 tokens. The corresponding semantic tree for $K=2$ is shown in Fig.~\ref{fig:example_reddit_story}. 

\begin{quote}
    Once there was a little girl who found a hidden band. She was so excited and took it with her to show her friends. When she showed it to her friends, they all wanted to play with it. But when the band got too close to the sun, it started to melt! All of the kids were so sad because the band was gone. But then one of them had an idea. They used their markers and colored paper to make a new band. It wasn't as nice as the first, but it was still fun to play with! The little girl was so glad that they found a way to make a new band and they still had lots of fun!
\end{quote}

\begin{figure*}
    \centering
    \includegraphics[width=1\textwidth]{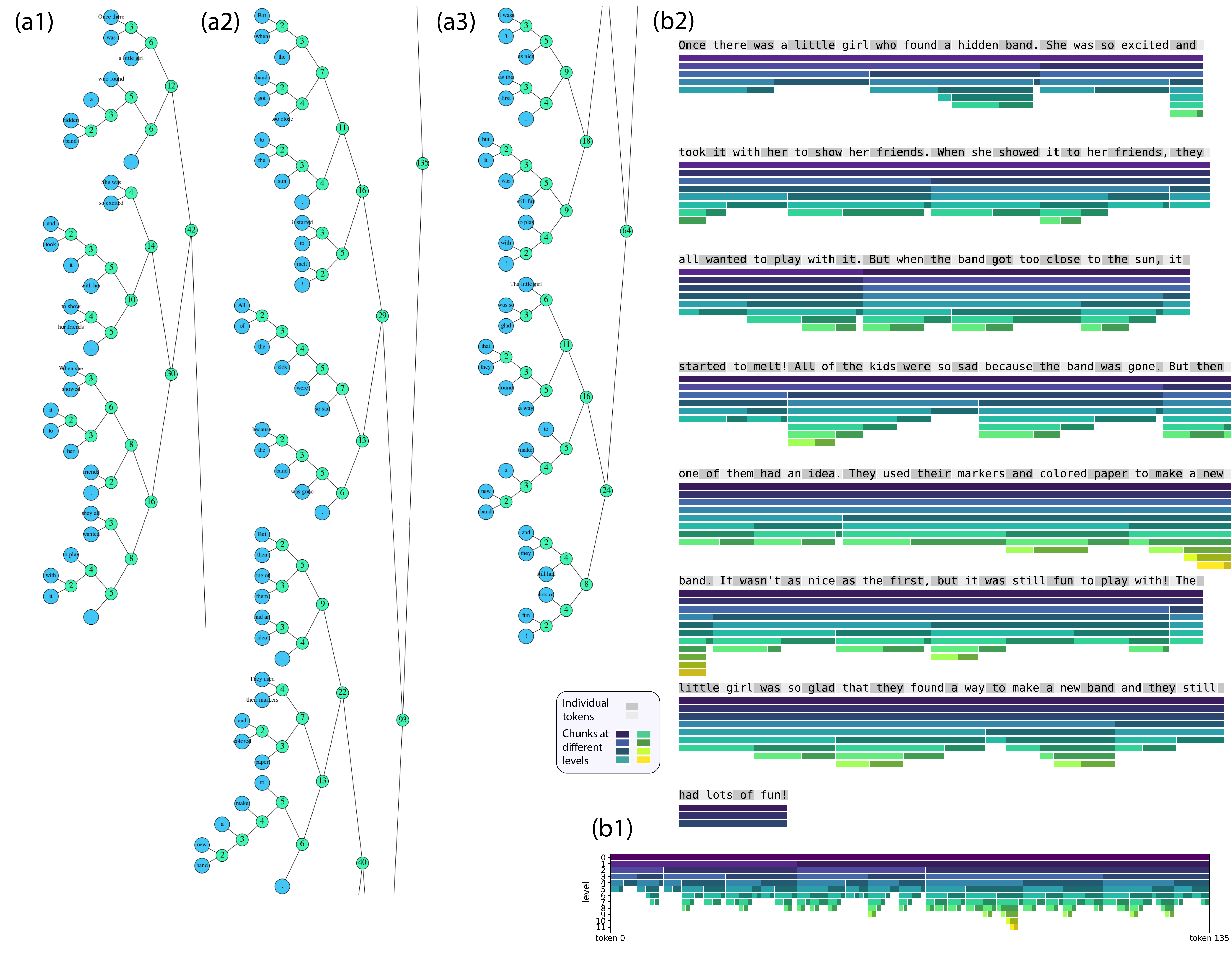}
    \caption{\textbf{Semantic tree of Tiny Story (198810).} The tree was originally plotted as a long vertical figure and is adapted here into three panels. Read the tree from top to bottom and left to right across panels (a)--(c). The root appears in panel (b) as the green node (135).}
    \vspace{-1em}
\label{fig:example_tiny_story}
\end{figure*}

\subsubsection*{Example \texttt{PoetryFoundationPoems} (8443)}

\begin{figure*}
    \centering
    \includegraphics[width=1\textwidth]{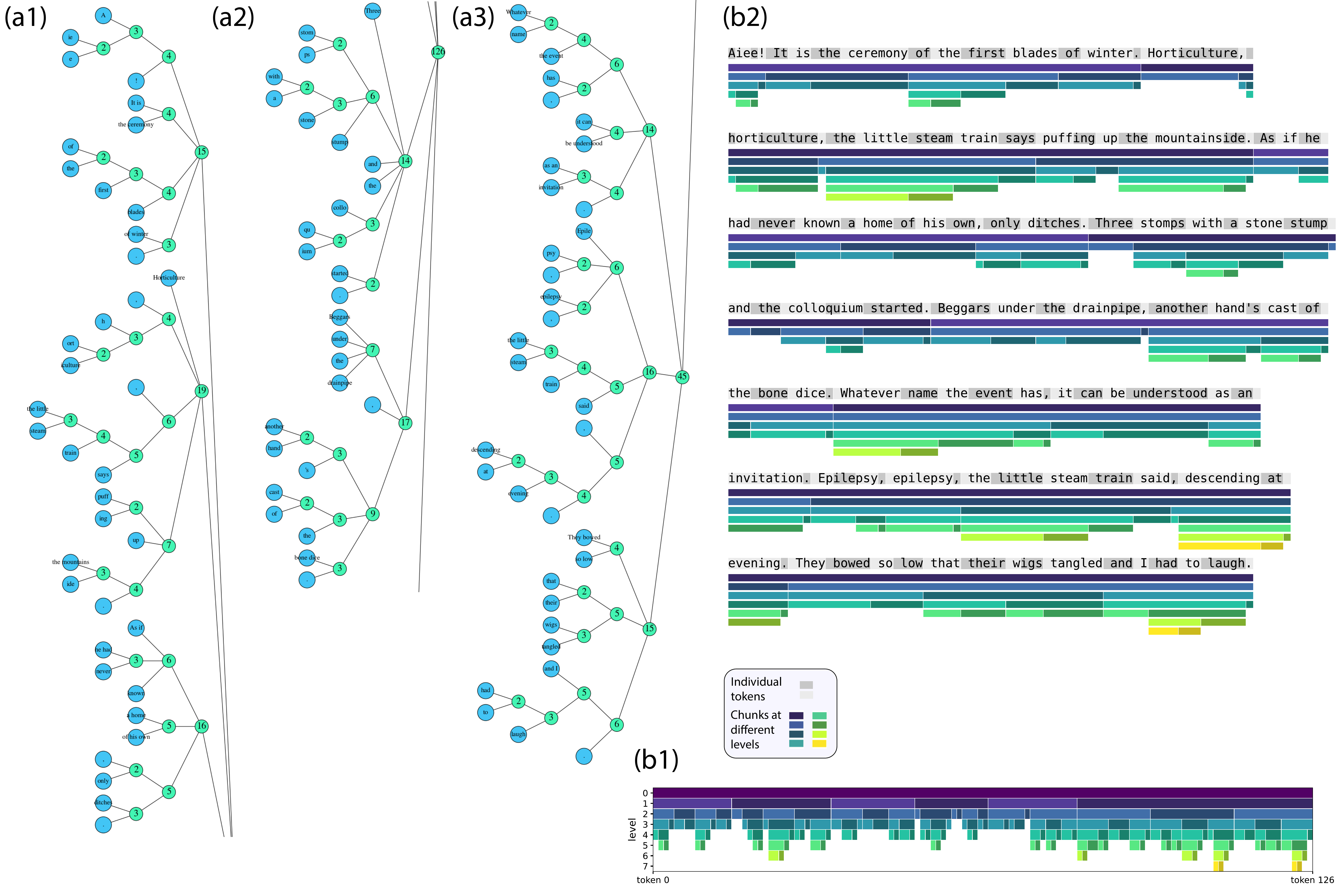}
    \caption{\textbf{Semantic tree of Modern Poetry (8443).} The tree was originally plotted as a long vertical figure and is adapted here into three panels. Read the tree from top to bottom and left to right across panels (a)--(c). The root appears in panel (b) as the green node (135).}
    \vspace{-1em}
\label{fig:example_modern_poetry}
\end{figure*}

In the following we give an example from the Modern Poetry (8443), it has a total of 126 tokens. The corresponding semantic tree for $K=6$ is shown in Fig.~\ref{fig:example_modern_poetry}. 

\begin{quote}
    Aiee! It is the ceremony of the first blades of winter. Horticulture, horticulture, the little steam train says puffing up the mountainside. As if he had never known a home of his own, only ditches. Three stomps with a stone stump and the colloquium started. Beggars under the drainpipe, another hand's cast of the bone dice. Whatever name the event has, it can be understood as an invitation. Epilepsy, epilepsy, the little steam train said, descending at evening. They bowed so low that their wigs tangled and I had to laugh.
\end{quote}

\subsection{Consistency across different models}

\begin{figure*}
    \centering
    \includegraphics[width=1\textwidth]{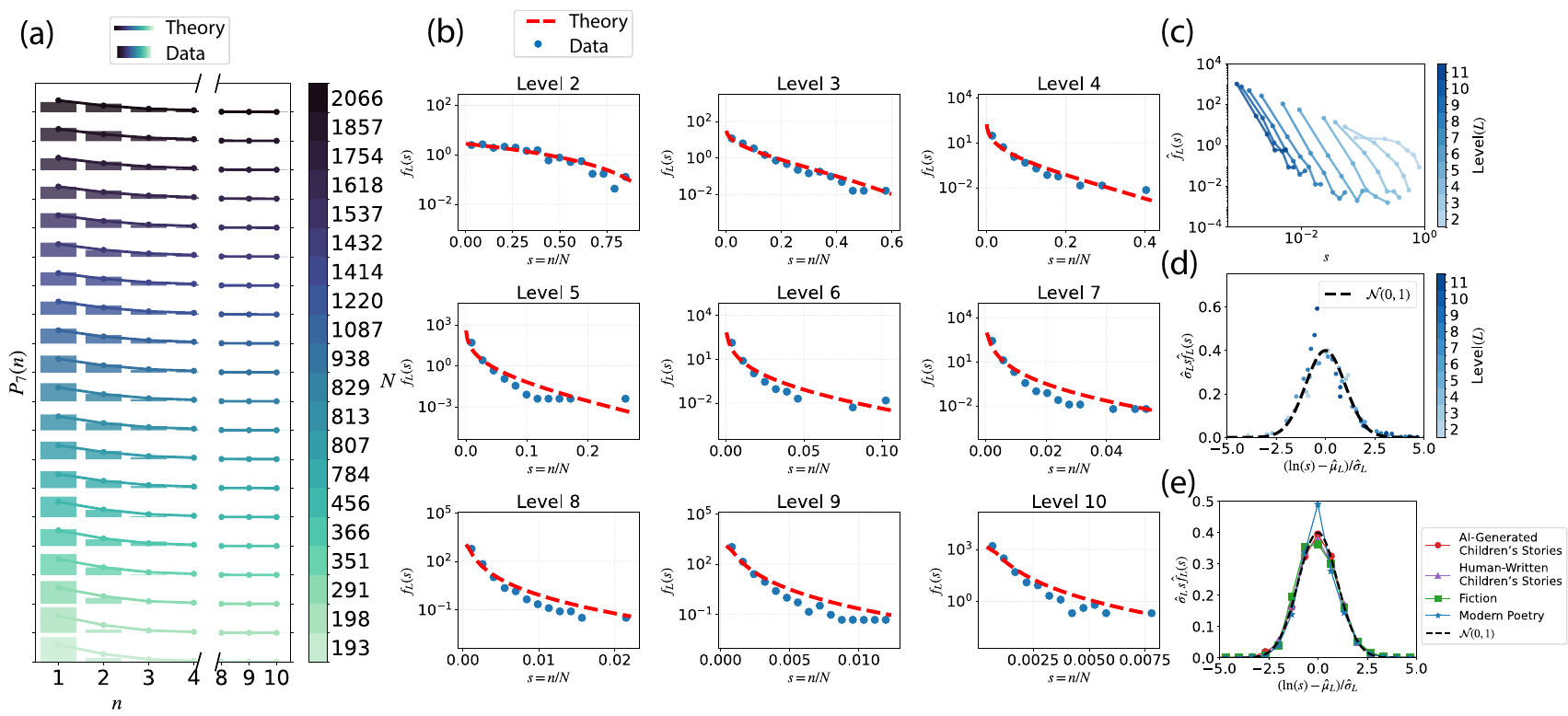}
    \caption{\textbf{Semantic chunking using \texttt{DeepSeek-V3.1}. } (a) Comparison between the theoretical and empirical chunk size distributions at an intermediate level (7) for 20 randomly selected texts from \texttt{RedditStories} of different lengths. (b) Comparison between the theoretical and empirical normalized chunk size distributions (pooled over 100 randomly selected texts from \texttt{RedditStories}) for level 3-11. (c) The same normalized chunk size distribution at differrent levels as in (b) but shown on a log-log plot. (d) The normalized chunk size distribution at different levels collapse to a universal standard normal distribution upon the transformation $x=(\ln s-\hat{\mu}_L)/\hat{\sigma}_L$. (e) The same data collapse is observed for different corpora. }
    \vspace{-1em}
\label{fig:deepseek}
\end{figure*}

In the following we present the results of semantic chunking using a different LLM, \texttt{DeepSeek-V3.1}, as opposed to \texttt{Llama-4-Maverick} as in main text Fig.~\ref{fig:tree_statistics} and Fig.~\ref{fig:tree_universal}. Fig.~\ref{fig:deepseek}(a) shows 20 randomly selected stories from the \texttt{RedditStories} dataset \cite{fan2018hierarchical}, where the empirical chunk size distributions at an intermediate level (7) is compared with the theoretical chunk size distributions and good agreement is observed. Fig.~\ref{fig:deepseek}(b) shows the comparison between empirical and theoretical normalized chunk size distribution (pooled over 100 randomly selected texts from \texttt{RedditStories}). Fig.~\ref{fig:deepseek}(c) shows the empirical (normalized) chunk size distributions in a log-log plot, and Fig.~\ref{fig:deepseek}(d) shows the data collapse as in main text Fig.~\ref{fig:tree_universal}(d). Fig.~\ref{fig:deepseek}(e) shows the same data collapse as in Fig.~\ref{fig:deepseek}(d) for three different datasets: Children's stories (\texttt{TinyStories}\cite{eldan2023tinystories}), fiction (\texttt{\texttt{RedditStories}}), and modern poetry (\texttt{PoetryFoundationPoems}\cite{hf_suayptalha_poetry_foundation_poems}). 

\section{Comparison between semantic and random chunking}

The agreement between the random tree theory and the empirical chunk-size distributions in main text Fig.~\ref{fig:tree_statistics} and SI Fig.~\ref{fig:deepseek} admits two possible interpretations. First, although each individual text is segmented according to semantic coherence, the aggregate statistics over a large corpus may be well described by a random tree ensemble, as proposed in the main text. Second, the chunking algorithm itself might ignore semantic structure and produce effectively random partitions, in which case agreement with a random model would be trivial. To distinguish these interpretations, we compare semantic chunking with an explicit random-chunking control, examining both chunk-size statistics and the preservation of linguistic clauses.

\begin{figure*}[t!]
    \centering
    \includegraphics[width=0.85\textwidth]{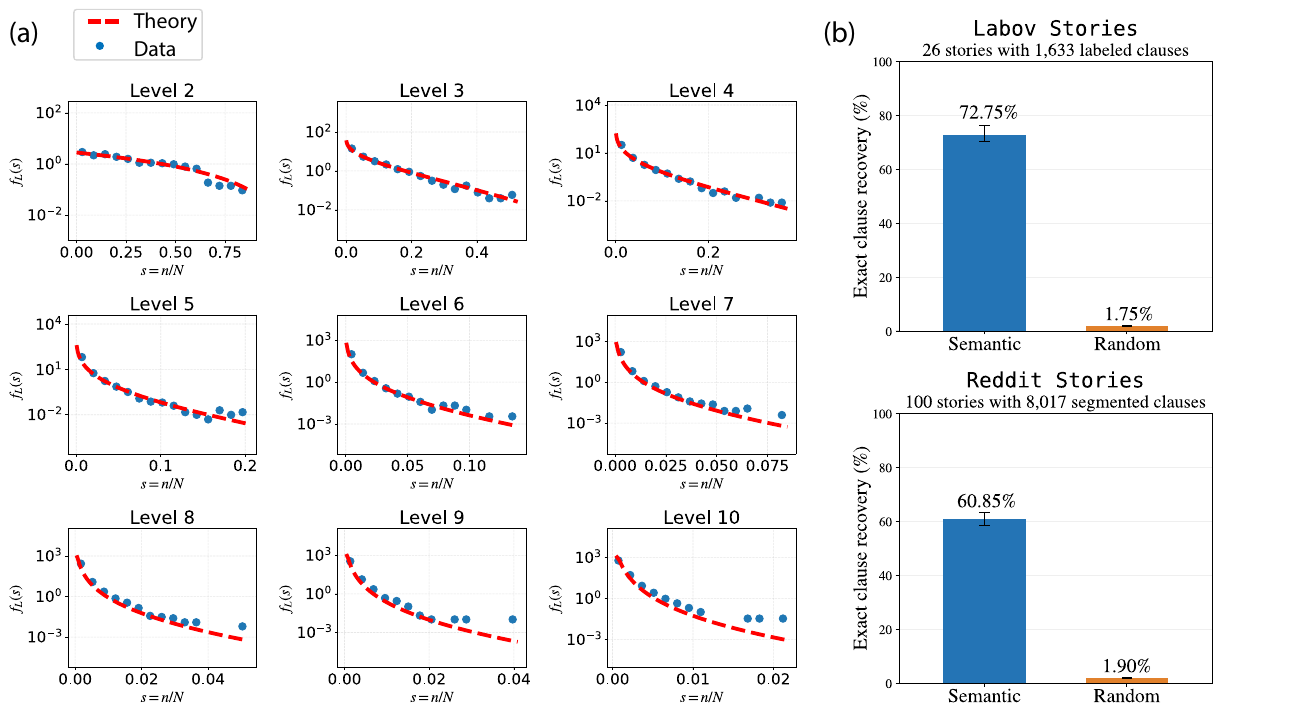}
    \caption{\textbf{Similar chunk-size statistics coexist with distinct linguistic structure.}
    \textbf{(a)} Random chunking produces normalized chunk-size distributions similar to those of semantic chunking [main text Fig.~\ref{fig:tree_statistics}(b)].
    \textbf{(b)} Exact clause recovery by semantic and random chunking. A clause is recovered when its complete span matches a single internal node at any tree level.
    Top: 26 \texttt{Labov Stories} with 1,633 labeled clauses~\cite{labov2013language}.
    Bottom: 100 \texttt{Reddit Stories} with 8,017 clauses segmented by \texttt{GPT-5.4-mini}, independently of the semantic trees generated using \texttt{Llama-4-Maverick}.
    Recovery fractions are pooled over clauses; random-tree results are averaged over 100 realizations per story.
    Error bars indicate 95\% confidence intervals obtained by resampling stories.
    }
    \vspace{-1em}
    \label{fig:random_chunking}
\end{figure*}

For each text, we construct a random tree using the same weak ordered-partition process as in the random tree model. Specifically, we treat a text as a sequence of $N$ tokens, or ``stars,'' together with $K-1$ chunk boundaries, or ``bars.'' The $K-1$ boundaries are placed uniformly at random among the $N+K-1$ possible stars and bars positions, producing $K$ possibly empty chunks. Adjacent boundaries correspond to empty chunks. We apply this partitioning procedure recursively to every non-empty child chunk, stopping a branch when a split leaves a child with the same size as its parent or when the chunk contains at most one token. We generate 100 independent random-tree realizations per story, using the same text and tokenization as for semantic chunking, and average their clause-recovery fractions. By construction, these simulations sample the random tree ensemble underlying the theoretical chunk-size distributions~\cite{stanley2012enumerative1,zhong2025random}. Their empirical distributions agree with the theoretical predictions (Fig.~\ref{fig:random_chunking}(a)), with residual deviations arising primarily from finite sampling, as discussed in Section~\ref{sec:deviations}.

We next test whether semantic and random trees differ in their preservation of linguistic structure. We first apply semantic chunking to 26 stories from \cite{labov2013language}, which include Labov's segmentation of the narratives into linguistic clauses. A clause is counted as recovered if its full span exactly matches a single non-root internal node at any level of the tree, retaining punctuation and ignoring only surrounding whitespace. Semantic chunking recovers 72.75\% of the 1,633 labeled clauses in the \texttt{Labov Stories}, whereas random chunking recovers only 1.75\% (top panel of Fig.~\ref{fig:random_chunking}(b)). We repeat this comparison on 100 \texttt{Reddit Stories}, for which clause segmentation are generated using \texttt{GPT-5.4-mini} \cite{openai2026gpt54mini}. We deliberately use a different model from the \texttt{Llama-4-Maverick} model used for semantic chunking to reduce potential bias from using the same model for both tasks. The clause-segmentation prompt is calibrated against the Labov annotations and achieves 86.4\% consistency on the \texttt{Labov stories}. In the resulting 8,017 segmented clauses, semantic chunking recovers 60.85\%, compared with 1.90\% for random chunking (bottom panel of Fig.~\ref{fig:random_chunking}(b)). Thus, despite their similar aggregate chunk-size statistics, semantic trees exhibit a strong preference for linguistic clause boundaries that is absent from the random control. This demonstrates that agreement with random tree model does not imply random partitioning of individual texts, and supports the interpretation that semantic chunking preserves semantically coherent units.

\vspace{-1em}
\section{Deviation from Randomness}
\label{sec:deviations}

In this section, we discuss the main sources of deviation between the empirical normalized chunk-size distributions \(\hat{f}_l(s)\) and the random tree ensemble scaling functions \(f_l(s)\) in Eq.~\eqref{eq:asymp_Pn}. For this study, we randomly choose 100 texts from \texttt{RedditStories}, which is a proper subset of the stories used in main text.

\begin{figure*}[ht]
\centering
\includegraphics[width=1\textwidth]{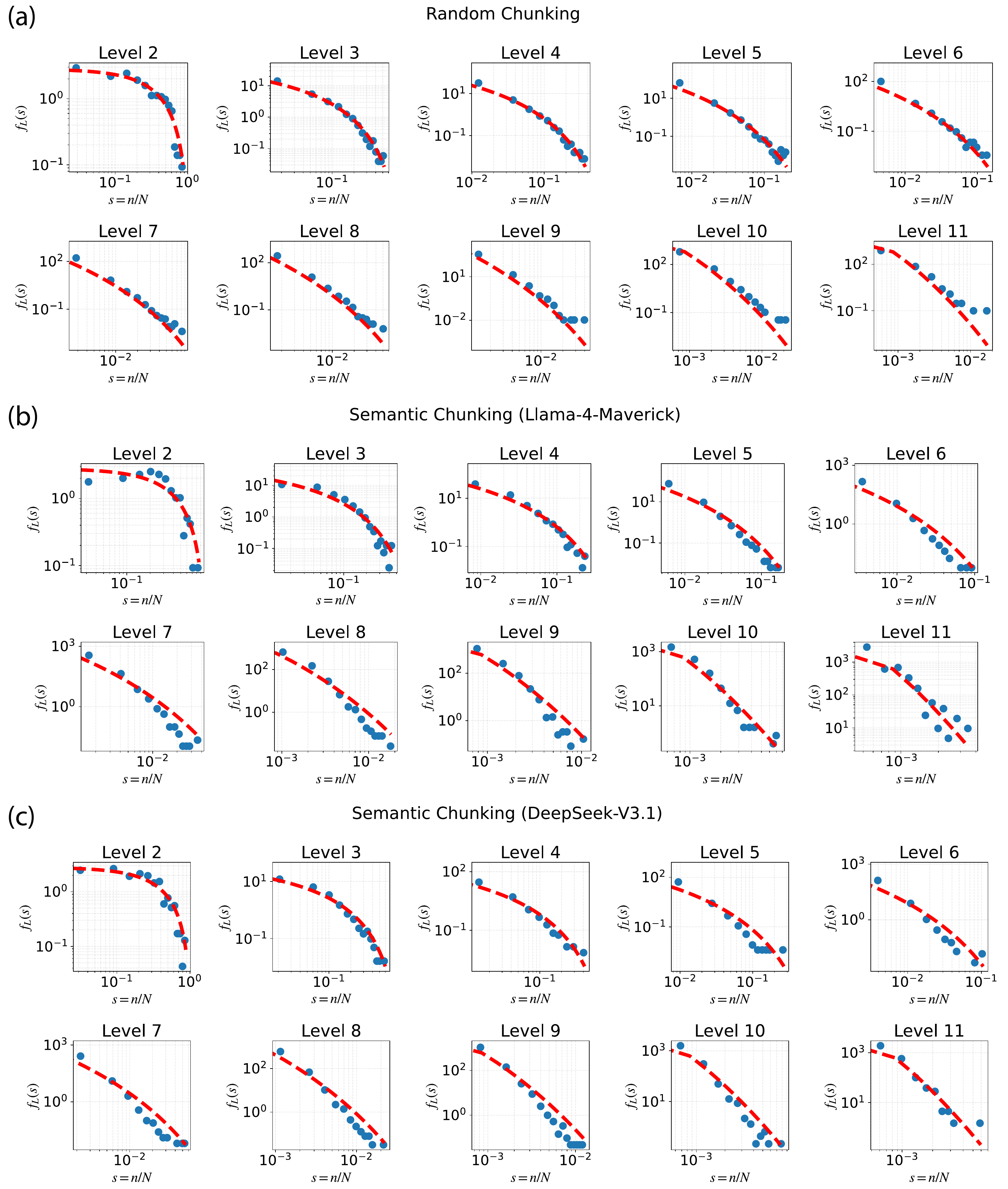}
\caption{\textbf{Robustness of random tree scaling across chunking procedures.}
Normalized chunk-size distributions are shown on log--log axes for \textbf{(a)} random chunking, \textbf{(b)} semantic chunking using \texttt{Llama-4-Maverick}, and \textbf{(c)} semantic chunking using \texttt{DeepSeek-V3.1}.}
\vspace{-1em}
\label{fig:deviations}
\end{figure*}

The first source of deviation is finite-size effects. The scaling function \(f_l(s)\) is an asymptotic prediction, so deviations are expected even for trees generated directly by the random chunking process. This is shown in Fig.~\ref{fig:deviations}(a), where the normalized chunk-size distributions from random chunking are plotted on log--log axes. Because finite-size effects are strongest for short documents, they are most visible at small normalized chunk sizes. For example, the first data points at many levels in Fig.~\ref{fig:deviations}(a)--(c) lie slightly above the theoretical curves. This deviation is the cost of pooling statistics across documents of heterogeneous lengths. In practice, the effect is modest because longer documents contribute more chunks to the pooled distributions.

A second source of deviation is finite-sample noise. Even if the random tree ensemble were an exact description, empirical estimates of \(\hat{f}_l(s)\) would fluctuate when the number of available chunks at a given level is small. This is most apparent near the shallowest and deepest levels, such as \(l=2\) and \(l=11\), where sample sizes are limited and deviations from the theoretical curves become more pronounced. These fluctuations can also depend on the chunking model, as seen by comparing the \texttt{Llama-4-Maverick} and \texttt{DeepSeek-V3.1} results in Fig.~\ref{fig:deviations}(b),(c), and may also depend on prompting details. We expect this source of deviation to decrease with larger corpora and to primarily affect the initial and final levels of the tree.

The third source of deviation is genuine non-randomness induced by semantic and linguistic constraints. At sufficiently deep levels, chunk sizes approach the clause or phrase scale, on the order of \(\sim 10\) tokens, where segmentation is strongly constrained by syntax, grammar, and semantic coherence. Empirically, this appears as an excess of small chunks, contributing to the upward shift of the first data points in Fig.~\ref{fig:deviations}(b),(c). Such deviations are expected: semantic chunking is not intended to produce purely random partitions, but rather to preserve coherent linguistic units. A systematic analysis of how this excess of small semantic chunks affects other quantities, such as the entropy of the tree ensemble, is beyond the scope of the present work. We leave this question to future work \cite{sivan2026}, where the entropy of random and non-random tree ensembles will be analyzed in greater detail.

\clearpage 

\section{Renormalization of chunk size distribution}

After taking the large $N$ limit, the normalized chunk size distribution is
\begin{align}
\label{eq:beta_kernel}
f_L(s) &= \int_s^1 \rho(s|s') f_{L-1}(s') ds'  \\
&= (K-1) \int_s^1 \frac{ds'}{s'} (1-\frac{s}{s'})^{K-2} f_{L-1}(s') ds'.
\end{align}
Let
\begin{align}
r = \frac{s}{s'}, \qquad ds' = -\frac{s'}{r} dr,    
\end{align}
then
\begin{align}
    f_L(s) = (K-1) \int_s^1 \frac{dr}{r} (1-r)^{K-2} f_{L-1}(\frac{s}{r}).
\end{align}
We can identify the Beta-distribution
\begin{align}
    p_B(r) = (K-1)(1-r)^{K-2} = \text{Beta}(1,K-1),
\end{align}
then
\begin{align}
\label{eq:Mellin_conv}
    f_L(s) = \int_s^1 \frac{dr}{r} p_B(r) f_{L-1}(\frac{s}{r}),
\end{align}
takes the form of a Mellin convolution, which describes the pdf of a product of two random variables. $f_L = p_B * f_{L-1}$. Therefore, the random variable for level-$L$ is a product of $(L-1)$ i.i.d. Beta-random variables 
\begin{align}
\label{eq:s_product}
s = r_1 r_2 \cdots r_{L-1}, \;\; r_i \sim p_B.   
\end{align}
For $L\to \infty$, $s$ is a product of many i.i.d. random variables, and we can use the central limit theorem in logspace: $f_L(s)$ becomes a lognormal distribution. Let
\begin{align}
    t_i &= \ln r_i = \frac{s_{i+1}}{s_i}, \\
    \ln s &= \sum_{i=1}^{L-1} t_i,
\end{align}
then $t_i$ is a log-transformed Beta random variable with mean and variance
\begin{align}
\label{eq:mu_sigma}
    \mu &= - \sum_{n=1}^{K-1}\frac{1}{n} = -H_{K-1}, \\
    \sigma^2 &= \sum_{n=1}^{K-1}\frac{1}{n^2} = H^{(2)}_{K-1},   
\end{align}
Then for the sum of $(L-1)$ such random variables, the distribution is
\begin{align}
\label{eq:lognormal_dist}
    f_L(s) &\simeq \frac{1}{\sqrt{2\pi}\sigma_L s} \exp \left\{ \frac{(\ln s - \mu_L)^2}{2\sigma_L^2} \right\}, \\
    \mu_L &= -(L-1)H_{K-1}, \\
    \sigma^2_L &= (L-1)H^{(2)}_{K-1}. 
\end{align}

\subsection{Sanity check: Compare lognormal with the exact solution for K=2}

Let's check the special case of $K=2$, which admits exact solution (Sec.~\ref{sec: K=2_exact})
\begin{align}
    f_L(s) = \frac{(-\ln s)^{L-2}}{(L-2)!}.
\end{align}
Let $x = -\ln s$, also assume $L=L-2 \gg 1 $, then (with Jacobian factor $e^{-x}$)
\begin{align}
    g_L(x) = f_L(s) e^{-x} = \frac{1}{L!} e^{ L\ln x -x}
\end{align}
Define $x:= L + \delta$, where we assume $\delta \sim O(\sqrt{L})$, we can rewrite
\begin{align}
    \ln x &= \ln (\delta + L) \\
    &= \ln L + \ln (1+\frac{\delta}{L}) \\
    &= \ln L + \frac{\delta}{L} - \frac{\delta^2}{2L^2} + O(L^{-3})
\end{align}
Then
\begin{align}
    \ln g_L(x) &= -\ln L! + L\ln L + \delta - \frac{\delta^2}{2L} - x \\
    &= L + \delta - \frac{\delta^2}{2L} - x - \frac{1}{2}\ln (2\pi L) \\
    &= - \frac{\delta^2}{2L} - \frac{1}{2}\ln (2\pi L),
\end{align}
where in the second line we have used the Stirling approximation 
\begin{align}
    \ln L! = L\ln L - L + \frac{1}{2}\ln (2\pi L) + O(\frac{1}{L}).
\end{align}
Therefore,
\begin{align}
    g_L(x) = \frac{1}{\sqrt{2\pi L}} \exp \left\{- \frac{(x-L)^2}{2L} \right\}.
\end{align}
Going back to $s = e^{-x}$, we have (for $K=2$)
\begin{align}
    f_L(s) &= \frac{1}{\sqrt{2\pi (L-2)}s} \exp \left\{- \frac{[\ln s + (L-2)]^2}{2(L-2)} \right\}, \\
\end{align}
which for $L \gg1 $ becomes identical to the $K=2$ case in Eq.\eqref{eq:lognormal_dist}, as $\mu_L = -(L-1)H_1 = -(L-1)$ and $\sigma_L^2 = (L-1)H^{(2)}_1 = (L-1)$. 

\subsection{RG flow of non-gaussianity}

The normalized chunk size variable $s$ becomes a lognormal random variable for $L \gg 1$, we define the rescaled $O(1)$ random variable 
\begin{align}
\label{eq:x_variable}
    x = \frac{\ln s - \mu_L}{\sigma_L}.
\end{align}
Let's consider the effect of adding one level, $L \to L+1$. 
\begin{align}
    s_{L+1} = r_L s_L, \;\; r_L \sim p_B = \text{Beta}(1,K-1),
\end{align}
So using the definitions in Eq.~\eqref{eq:mu_sigma} and Eq.~\eqref{eq:x_variable}, we can write
\begin{align}
\label{eq:D_step_x}
    x_{L+1} &= \frac{\ln s_{L+1} - L\mu }{\sqrt{L}\sigma} \\
    &= \sqrt{\frac{L-1}{L}} x_L + \frac{1}{\sqrt{L}} \frac{\ln r_L - \mu}{\sigma} \\
    &:= a_L x + b_L y,
\end{align}
where we have isolated an $O(1)$ and an $O(L^{-1/2})$ contribution
\begin{align}
   a_L &= 1-\frac{1}{2D} + O(L^{-2}), \\
   b_L &= \frac{1}{\sqrt{L}}.
\end{align}
The log-variables are distributed as
\begin{align}
\label{eq:log_var_dists}
    x &\sim \sigma\sqrt{L-1}sf_L(s):= P_L(x), \\
    y &\sim \sigma s f_2(s) := P_2(y), \nonumber
\end{align}
where $P_2(y)$ is the pdf for the 1-step log-beta variable $y=\ln r_L, \; r_L \sim p_B$, and $P_L(x)$ is the pdf of the $(L-1)$ step log-variable in Eq.~\eqref{eq:x_variable}. 

Therefore, the $L$ step log-variable (Eq.~\eqref{eq:D_step_x}) has pdf 
\begin{align}
    P_{L+1}(z) &= \int_{-\infty}^{\infty} P_L(x) P_2(y) \delta(z-a_L x - b_L y) dx dy \\
     &= \frac{1}{a_L} \int_{-\infty}^{\infty} P_L\left(\frac{z-b_L x}{a_L}\right) P_2(x) dx \label{eq:RG_map_real}
\end{align}
Fourier transforming Eq.~\eqref{eq:RG_map_real} we obtain
\begin{align}
\label{eq:RG_map_fourier}
    \hat{P}_{L+1}(k) = \hat{P}_L(a_L k) \hat{P}_2(b_Lk).
\end{align}
Eq.~\eqref{eq:RG_map_real} is the ``real-space'' RG map transforming the pdf from $L$ to $L+1$, while Eq.~\eqref{eq:RG_map_fourier} is the Fourier-space RG map.

By Central Limit Theorem (CLT), $P_L(z)$ becomes a normal distribution for $L\gg 1$, so let's try to expand around this Gaussian fixed point,
\begin{align}
    P_L(z) &= \phi(z) \left[ 1 + \sum_{m \geq 0}^{\infty} \frac{g_m(L)}{m!} H_m(z) \right], \\
    \phi(z) &= \mathcal{N}(0,1),
\end{align}
where $g_m(L)$ are the coefficients of the non-gaussianity modes (not all modes are free to vary, see below), and $H_m(z)$ are the Hermite polynomials satisfying the following properties:
\begin{align}
    H_m(z)&:= (-1)^m e^{\frac{z^2}{2}} \partial_z^m e^{-\frac{z^2}{2}}, \\
    H_m(z)\phi(z) &= (-1)^m \partial_z^m \phi(z), \\
    n!\delta_{mn}  &= \int_{-\infty}^{\infty} dz \phi(z) H_m(z) H_n(z). \label{eq:orthogonality}
\end{align}
Now we impose the constraints that $P_L(z)$ needs to be normalized, and has zero mean, unit variance:
\begin{align}
    1 &= \int_{-\infty}^{\infty} dz P_L(z) = \int_{-\infty}^{\infty} dz P_L(z)H_0(z) \\
    0 &= \int_{-\infty}^{\infty} dz P_L(z)z = \int_{-\infty}^{\infty} dz P_L(z)H_1(z) \\
    1 &= \int_{-\infty}^{\infty} dz P_L(z) z^2 = \int_{-\infty}^{\infty} dz P_L(z) [1+H_2(z)],
\end{align}
where we have used the fact that the first three Hermite polynomials are:
\begin{align}
    H_0(z) &= 1 \\
    H_1(z) &= z  \\
    H_2(z) &= z^2 -1.
\end{align}
Using the orthogonality condition of the Hermite polynomials (Eq.~\eqref{eq:orthogonality}), we have $g_0 = g_1 = g_2 = 0$.

Next, since 
\begin{align}
    \mathcal{F}[\phi](k) &:= \hat{\phi}(k) = e^{-\frac{k^2}{2}}, \\
    \mathcal{F}[H_m\phi] &= (-1)^m \mathcal{F}[\partial_z^m \phi(z)] = (ik)^m \hat{\phi}(k),
\end{align}
The Fourier transformed pdf becomes
\begin{align}
    \hat{P}_L(k) = e^{-\frac{k^2}{2}} \left[1 +  \sum_{m \geq 3} g_m \frac{(ik)^m}{m!} \right].
\end{align}
Let's define
\begin{align}
    \Phi_L(k) &= \ln \hat{P}_L(k) = \ln \mathbb{E}\left[e^{ikz}\right] \\
    &= -\frac{k^2}{2} + \ln \left[1 + \sum_{m \geq 3} g_m \frac{(ik)^m}{m!} \right] \\
    &\approx -\frac{k^2}{2} + \sum_{m\geq 3} g_m(L) \frac{(ik)^m}{m!}, \label{eq:cumulant_expansion}
\end{align}
then it becomes apparent that $g_m(L)$ are the cumulants of $P_L(z)$.

Now the goal is to express both sides of Eq.~\eqref{eq:RG_map_fourier} in terms of these cumulant expansions, and match the coefficients. The RHS is:
\begin{align}
    \Phi_L(a_Lk) &= -\frac{1}{2}a_L^2k^2 + \sum_{m} g_m(L) a_L^m \frac{(ik)^m}{m!}, \\
    \Phi_2(b_Lk) &= -\frac{1}{2}b_L^2k^2 + \sum_{m} \gamma_m(L) b_L^m \frac{(ik)^m}{m!}.
\end{align}
Therefore (noting that $a_L^2 + b_L^2 = 1$)
\begin{align}
\Phi_{L+1}(k) 
&= -\frac{1}{2}k^2 + \sum_{m} \left[ a_L^m g_m(L) + b_L^m \gamma_m(L) \right] \frac{(ik)^m}{m!} \\
&= -\frac{1}{2}k^2 + \sum_{m} g_m(L+1) \frac{(ik)^m}{m!},
\end{align}
we can read off the RG update of the coefficients:
\begin{align}
\label{eq:gm_ab}
    g_m(L+1) = a_L^m g_m(L) + b_L^m \gamma_m(L).
\end{align}
Next, we relate $g_m$ with $\gamma_m$, using the fact that they come from the same underlying family of distributions $f_L(s)$ (Eq.~\eqref{eq:log_var_dists}). Let's measure everything in terms of the cumulant of the unscaled zero-mean log-variable $(\ln r_i -\mu) $,  $\kappa_m$:
\begin{align}
    \gamma_m[y] = \gamma_m \left[\frac{\ln r_i - \mu}{\sigma}\right] = \frac{\kappa_m}{\sigma^m},
\end{align}
whereas
\begin{align}
    g_m[x] &= g_m\left[ \frac{\sum_{i}^{L-1} \ln r_i - (L-1)\mu}{\sqrt{L-1}\sigma} \right] \\
    &= \frac{(L-1)\kappa_m}{(L-1)^{m/2}\sigma^m} \\
    &= \frac{\gamma_m}{(L-1)^{\frac{m-2}{2}}}.\label{eq:gamma_g_relation}
\end{align}
Therefore, inserting Eq.~\eqref{eq:gamma_g_relation} back to Eq.~\eqref{eq:gm_ab} we have
\begin{align}
    g_m(L+1) &= g_m(L) \left[ a_L^m + b_L^m (L-1)^{\frac{m}{2}-1} \right] \\
    &= g_m(L) \left[1- \frac{m-2}{2D}  + O(L^{-2})  \right]. \label{eq:discrete_update}
\end{align}
We can turn Eq.~\eqref{eq:discrete_update} into a differential equation
\begin{align}
\label{eq:continuous_update}
    \frac{d \ln g_m(L)}{dD} &= -\frac{m-2}{2} \frac{1}{L}.
\end{align}
Solving Eq.~\eqref{eq:continuous_update} gives
\begin{align}
\label{eq:scaling_dimension}
    g_m(L) &= L^{-\Delta_m}, \\
    \Delta_m &= \frac{m-2}{2}, \;\; m\geq 3 ,
\end{align}
where $\Delta_m$ are the scaling dimensions of the non-gaussianity fluctuation modes. 
Following the RG conventions, the beta-function of the RG flow is (defining $\ell = \ln L$)
\begin{align}
    \frac{dg_m(\ell)}{d\ell} &= -\frac{m-2}{2} g_m(\ell).
\end{align}
Since $g_m(L)$ are the cumulants of $P_L$ (Eq.~\eqref{eq:cumulant_expansion}), Eq.~\eqref{eq:scaling_dimension} confirms that the CLT we have assumed is self-consistent: all $m \geq 3$ cumulants are irrelavant directions of the RG flow, and all these higher cumulants vanishes asymptotically as $L \to \infty$. 

This derivation can be thought of as a RG derivation of the central limit theorem, in a spirit similar to the treatment in \cite{jona_lasinio2001rgprobability} for continuous RG transformations, but with important distinction that this applies to discrete random $K$-ary trees.

\end{document}